\documentclass{article}

\usepackage[final]{corl_2022} 
\usepackage{amsmath}
\usepackage{mathtools}
\usepackage{amsfonts}
\usepackage{bm}
\usepackage{color}
\usepackage{xcolor}
\usepackage{amsthm}
\usepackage{algpseudocode}
\usepackage{algorithm}
\usepackage{multirow}
\usepackage{subfigure}
\usepackage{booktabs}
\usepackage{enumitem}
\usepackage{multirow}
\usepackage{graphicx}
\usepackage{subfigure}
\usepackage{wrapfig}
\usepackage{siunitx}

\DeclareMathOperator*{\argmax}{arg max}
\DeclareMathOperator*{\argmin}{arg min}
\usepackage{mathtools}

\DeclarePairedDelimiter\floor{\lfloor}{\rfloor}
\newtheorem{theorem}{Theorem}
\newtheorem{definition}{Definition}
\newtheorem{lemma}{Lemma}

\newtheorem{corollary}{Corollary}
\newtheorem{remark}{Remark}

\theoremstyle{plain}

\theoremstyle{plain}

\theoremstyle{plain}

\theoremstyle{plain}

\theoremstyle{plain}

\definecolor{revise}{RGB}{0, 0, 0}

\newcommand{\xqedhere}[2]{%
\rlap{\hbox to#1{\hfil\llap{\ensuremath{#2}}}}}

\title{Robustness Certification of Visual Perception Models via Camera Motion Smoothing}

\author{Hanjiang Hu$^{1}$ Zuxin Liu$^{1}$\quad  Linyi Li$^{2}$\quad  Jiacheng Zhu$^{1}$ \quad Ding Zhao$^{1}$\\
$^1$Carnegie Mellon University\quad   $^2$University of Illinois at Urbana-Champaign
\\ {\tt\small \{hanjianghu,dingzhao\}@cmu.edu, \{zuxinl,jzhu4\}@andrew.cmu.edu, linyi2@illinois.edu}
}

\begin{document}
\maketitle


\begin{abstract}
A vast literature shows that the learning-based visual perception model is sensitive to adversarial noises but few works consider the robustness of robotic perception models under widely-existing camera motion perturbations.
    To this end, we study the robustness of the visual perception model under camera motion perturbations to investigate the influence of camera motion on robotic perception.
    Specifically, we propose a motion smoothing technique for arbitrary image classification models, whose robustness under camera motion perturbations could be certified.
    The proposed robustness certification framework based on camera motion smoothing provides effective and scalable robustness guarantees for visual perception modules so that they are applicable to wide robotic applications.
     As far as we are aware, this is the first work to provide the robustness certification for the deep perception module against camera motions, which improves the trustworthiness of robotic perception.
    A realistic indoor robotic dataset with the dense point cloud map for the entire room, \textit{MetaRoom}, is introduced for the challenging certifiable robust perception task.
    We conduct extensive experiments to validate the certification approach via motion smoothing against camera motion perturbations.
    Our framework guarantees the certified accuracy of 81.7\%  against camera translation perturbation along depth direction within -0.1m $\sim$ 0.1m. \textcolor{revise}{We also validate the effectiveness of our method on real-world robot by conducting hardware experiment on robotic arm with an eye-in-hand camera.} The code is available on \url{https://github.com/HanjiangHu/camera-motion-smoothing}.
\end{abstract}

\keywords{Certifiable  Robustness, Camera Motion Perturbation,  Robotic Perception}


\section{Introduction}

Visual perception has achieved great success in recent years by leveraging the powerful representation capability of neural networks and the diversity of large datasets~\cite{iandola2016squeezenet, hu2020seasondepth, deng2009imagenet}.
Deep learning models have been dominating a wide range of computer vision tasks, such as image classification \cite{he2016deep,dosovitskiy2020image}, object detection~\cite{redmon2016you, xu2022opv2v, xu2022v2x} and segmentation \cite{he2017mask, hu2020dasgil, xu2022cobevt}.
However, applying deep perception models to real-world robotic applications is still challenging.
Since visual perception is the core upstream module of an autonomous robot system, its failure can cause the robot to sense the surrounding environments incorrectly, which may result in catastrophic consequences.
Therefore, developing a trustworthy perception system that can guarantee functionality in diverse real-world scenarios is necessary \cite{eykholt2018robust, hu2021seasondepth, yang2022certifiably}.

\begin{figure}
    \centering
    \includegraphics[width=0.9\textwidth]{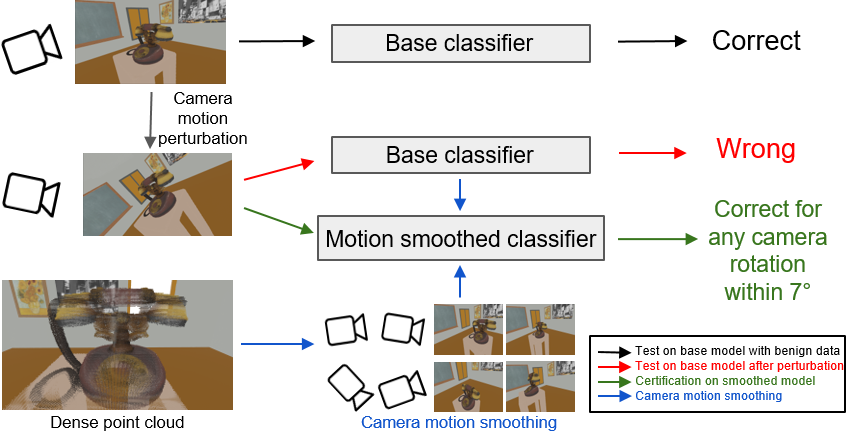}
    \vspace{-3mm}
    \caption{\small Certification framework via camera motion smoothing}
    \label{fig:overview}
        \vspace{-6mm}
\end{figure}

We study the robustness of a visual perception system  against sensing noise due to camera motion perturbation that commonly exists in robotic applications~\cite{ferreira2014probabilistic, liu2022robustness, yang2022certifiably}, which is important while challenging for  trustworthy robotic applications.
The difficulty arises from two perspectives: internal model vulnerability and external sensing uncertainty.
On the one hand, rich literature suggests that deep visual models are vulnerable to adversarial perturbations that would be stealthy to human eyes: small perturbations added to the input of neural networks can significantly corrupt the perception performance~\cite{goodfellow2014explaining, kurakin2018adversarial, carlini2017towards, xiao2018generating}.
On the other hand, several recent works indicate that the perception performance is also sensitive to data acquisition, such as sensor placement and sensing perspective~\cite{liu2019should, feng2021bridging, hu2022investigating, tang2017precision}.
Both internal vulnerability and external sensing uncertainty make it challenging to guarantee the robustness of a visual perception system in real-world robotic applications.

Prior works propose several techniques to improve the visual perception system robustness~\cite{tramer2018ensemble,ma2018characterizing, tramer2020adaptive}, though most of them are demonstrated to be effective empirically and no theoretical robustness guarantees are given.
A recent line of work aims to provide provable robustness certification or verification such that the perception system is guaranteed to function properly under any bounded adversarial perturbations~\cite{cohen2019certified, tjeng2018evaluating, wong2018provable} or semantic image transformations and deformations \cite{ li2021tss, ruoss2021efficient, alfarra2021deformrs}, which improves the trustworthiness of the perception model.
However, most of them are studied under static 2D image datasets without viewpoint changes, while the robotic visual perception systems process 2D images projected from the 3D physical world, which may not be robust under different perspectives of the moving camera and make it challenging to certify the robustness for real-world robotic applications.

To tackle the challenge, we first study the robustness of the robotic perception model against the camera motion perturbation.
Next, we propose the \emph{first} framework with a certified robustness guarantee for robotic perception models against arbitrary bounded camera motion perturbations based on a novel motion smoothing strategy as shown in Fig.~\ref{fig:overview}.
\textcolor{revise}{Extensive experiments have been conducted on a realistic indoor robotic dataset \textit{MetaRoom} with the dense point cloud map as an given oracle for image projection, which is collected from the Webots simulation environment \cite{Webots} to show the effectiveness of the proposed robustness certification method against camera motion perturbations. To the best of our knowledge, this is the first work to study the certifiable robustness of image-based robotic perception under camera ego-motion based on motion smoothing. The contributions are summarized as follows.}
\setlength{\itemsep}{0pt}
\setlength{\parsep}{0pt}
\setlength{\parskip}{0pt}
\begin{itemize}[leftmargin=*]
\item We demonstrate that the camera motion perturbation can significantly influence perception performance and introduce motion smoothing to improve robustness over camera motion perturbation.
\item We propose a smoothing algorithm for any black-box image classification model such that its robustness against camera motion perturbations can be certified by our certification framework.
\item We conduct extensive experiments on the realistic indoor robotic simulation to validate the effectiveness of the motion smoothing certification, achieving over 80\% certified accuracy against any perturbations within radii of 0.1m or 7$^\circ$ for camera translation or rotation along the depth axis. \textcolor{revise}{Further experiments on real-world robot arm validate the effectiveness of camera motion smoothing for perception models.}
\end{itemize}



\section{Related Work}
\textbf{Robust Robotic Perception.}
The robustness of robotic perception has been studied from different viewpoints. 
A rich literature in the robust machine learning community shows that deep learning-based robotic perception models are vulnerable and can be easily fooled by adversarial samples~\cite{goodfellow2014explaining, kurakin2018adversarial, carlini2017towards, xiao2018generating}. 
Another perspective for robotic perception is the external sensing uncertainty, which is caused by sensor placement \cite{liu2019should, hu2022investigating}, camera distortions \cite{tang2017precision}, geometric outliers \cite{yang2022certifiably}, long-term robustness \cite{hu2021domain, hu2019retrieval}, sim-to-real adaptation \cite{tobin2017domain, hu2020dasgil, qiao2021registration}, etc. 
However, the robustness of deep perception models given the perturbed projected images from 3D physical world with a moving camera sensor is relatively understudied. 

\textbf{Certifiable Robustness under Perturbations.}
In recent years, a significant number of approaches has been proposed to provide robustness certification for deep neural networks~\cite{li2020sok,liu2019algorithms}.
In contrast to empirical robustness approaches~\cite{madry2017towards,tramer2018ensemble,ma2018characterizing, tramer2020adaptive}, i.e., which train robust models against adversarial perturbations, the robustness certification approaches aim to guarantee the accuracy of the perception model as long as the perturbation magnitude is bounded by some threshold.
Such robustness certification approaches have been proposed against both $\ell_p$-bounded pixel-wise perturbations~\cite{cohen2019certified, tjeng2018evaluating, wong2018provable,zhang2018efficient,singh2019abstract,dathathri2020enabling} and semantic transformation or deformations~\cite{zhongkai2022gsmooth,li2021tss, ruoss2021efficient, alfarra2021deformrs,balunovic2019certifying}, providing either deterministic guarantees based on function relaxations~\cite{balunovic2019certifying, mohapatra2020towards, lorenz2021robustness, ruoss2021efficient} or high-confident probabilistic guarantees based on random smoothing~\cite{fischer2020certified, li2021tss, alfarra2021deformrs, chu2022tpc, zhongkai2022gsmooth}. However, they consider either 2D or 3D transformations.
To the best of our knowledge, no prior work studies certifiable robustness associated with the movement of sensors and 3D-2D projected images, despite the fact that it is commonly seen in trustworthy robotic applications. 
Therefore, we aim to bridge the camera motion perturbation with deep learning robustness certification for  perception systems.










 




\section{Methodology}
In this section, we introduce the robustness certification framework against camera motion perturbations through the motion-smoothed perception model. We first define the image projection in terms of camera motion. Then, we clarify the certification goal and define the camera motion smoothed classifier using image projection. Finally, we present the robustness certification for each decomposed translation and rotation translation.

\subsection{Image Projection with Camera Motion}
We first define the positive projection in Def. \ref{def:oracle} based on the camera imaging concept in geometric computer vision \cite{szeliski2010computer}. The follow-up Def. \ref{def:proj_trans} defines the relative projective transformation $\phi(x,\alpha)$ parameterized by relative camera motion $\alpha$ with respect to the image $x$ at motion origin.
\begin{definition}[Position projective function]
\label{def:oracle}
For any  3D point $P=(X, Y, Z)\in\mathbb{P}\subset\mathbb{R}^3$ under the camera coordinate frame with  the camera intrinsic matrix $K$, based on the camera motion $\alpha= (\bm{\theta}, t) \in \mathcal{Z}\subset\mathbb{R}^6$ with  rotation matrix $ R = \exp(\bm{\theta}^\wedge) \in SO(3)$ and translation vector $t\in \mathbb{R}^3$, we define the position projective function $\rho: \mathbb{P}\times\mathcal{Z}\to \mathbb{R}^2$  and the depth function $D: \mathbb{P}\times\mathcal{Z}\to \mathbb{R}$  for point $P$ as
    \begin{align}
    \centering
    \label{def:projection}
    [\rho(P, \alpha), 1]^\top = \frac{1}{D(P, \alpha)}KR^{-1}(P - t), \quad
    D(P, \alpha) = [0, 0, 1] R^{-1}(P - t)
    \end{align}
\end{definition}

\begin{definition}[Channel-wise projective transformation]
\label{def:proj_trans}
Given the position projection function $\rho: \mathbb{P}\times\mathcal{Z}\to \mathbb{R}^2$  and the depth function $D: \mathbb{P}\times \mathcal{Z}\to \mathbb{R}$ over dense 3D point cloud $\mathbb{P}$, define the  3D-2D global channel-wise projective transformation from $C$-channel colored point cloud $\mathbb{V}=(\mathbb{R}^C, \mathbb{P})\subset \mathbb{R}^{C+3}$ to $H\times W$ image gird $\mathcal{X}\subset \mathbb{R}^{C\times H\times W}$ as $O: \mathbb{V}\times\mathcal{Z} \to \mathcal{X}$ parameterized by camera motion $\alpha\in \mathcal{Z}$ using Floor function $\floor{\cdot}$,
\begin{align}
\label{def:min_pooling}
        x_{c,r,s}= O(V, \alpha)_{c,r,s} =         V_{c, P^*_\alpha}, \text{where } P^*_\alpha = \argmin \limits_{\{P\in \mathbb{P} \mid \floor{\rho(P,\alpha)} = (r,s) \}} D(P, \alpha)
\end{align}
Specifically, if $x = O(V, 0)$, we define the relative projective transformation  $\phi: \mathcal{X}\times\mathcal{Z}\to\mathcal{X}$ as,
\begin{align}
\phi(x,\alpha)= O(V,\alpha).
\end{align}
\end{definition}

\subsection{Certification Goal and Motion Smoothed Classifier}
\textbf{Certification Goal.} We consider the classification task as the most fundamental perception task. For any deep learning-based classification model $h$, given the projected image $x$ at the origin of camera motion in the motion space $\mathcal{Z}$, the certification goal is to find a set within a radius $\mathcal{Z}_{\mathrm{radius}}\subseteq\mathcal{Z}$ such that, \textcolor{revise}{for any relative projective transformation  $\phi \in \mathcal{Z}_{\mathrm{radius}} $, with high confidence we have}
    \begin{equation}
    \label{certification_goal}
        h(x) = h(\phi(x,\alpha)), \forall \alpha\in\mathcal{Z}_{\mathrm{radius}}.
    \end{equation}

Based on the relative projective transformation $\phi$ over camera motion space $\mathcal{Z}$ defined above, we define the camera motion smoothed classification model as follows.
\begin{definition}[Camera motion $\varepsilon$-smoothed classifier]
\label{def:smooth_parametric_classifier}
Let $\phi: \mathcal{X}\times\mathcal{Z}\to\mathcal{X}$ be a relative projective transformation given the projected image $x$ at the origin of camera motion in the motion space $\mathcal{Z}$,
 and let $\varepsilon\sim\mathcal{P}_\varepsilon$ be a random camera motion taking values in $\mathcal{Z}$. Let $h: \mathcal{X}\to\mathcal{Y}$ be a base classifier $h(x) = \argmax_{y\in\mathcal{Y}} p(y\mid x)$, the expectation of projected image predictions $\phi(x, \varepsilon)$ over  camera motion distribution $\mathcal{P}_\varepsilon$  is $q(y\mid x;\varepsilon) := \mathbb{E}_{\varepsilon\sim\mathcal{P}_\varepsilon}p(y\mid\phi(x,\varepsilon))$. We define the $\varepsilon$-smoothed classifier $g:\mathcal{X}\to\mathcal{Y}$ as
     \begin{equation}
        g(x;\varepsilon) := \argmax_{y\in\mathcal{Y}} q(y\mid x;\varepsilon) = \argmax_{y\in\mathcal{Y}} \mathbb{E}_{\varepsilon\sim\mathcal{P}_\varepsilon} p(y\mid\phi(x,\varepsilon)).
    \end{equation}
\end{definition}

\subsection{Certifying Motion-parameterized Projection with Smoothed Classifier}
\textcolor{revise}{In order to achieve the certification goal \ref{certification_goal} with smoothed classifier, prior works \cite{li2021tss, chu2022tpc, hao2022gsmooth} show that smoothed model can be certified if the image transformation is resolvable. To this end,  we first show that the relative projection is generally compatible with the global projection,  which indicates that image projection can be regarded as resolvable transformation.}
\begin{lemma}[Compatible Relative Projection with Global Projection]
\label{lemma:resolvable_2d_proj}
            With a global projective transformation $O: \mathbb{V}\times\mathcal{Z} \to \mathcal{X}$ from 3D point cloud and a relative  projective  transformation $\phi: \mathcal{X}\times\mathcal{Z}\to\mathcal{X}$ given some original camera motions,  for any $\alpha_1\in\mathcal{Z}$ there exists an injective, continuously differentiable and non-vanishing-Jacobian function $\gamma_{\alpha_1}:\mathcal{Z}\to\mathcal{Z}$  such that 
            \begin{equation}
            \label{lemma:gamma}
                \phi(O(V, \alpha_1),\alpha_2) = O(V,\gamma_{\alpha_1}(\alpha_2)), V\in\mathbb{V},\alpha_2\in\mathcal{Z}.
            \end{equation}
\end{lemma}
The high-level idea for the proof is that $SO(3)$ has the rules of multiplication and inverse operation which can be transferred to the $\gamma_{\alpha_1}$ function (referred as $\gamma$ for convenience) in \eqref{lemma:gamma}, and min-pooling defined in the projective transformation in \eqref{def:min_pooling} does not break these rules as well. The full proof can be found in the Appendix materials.
\textcolor{revise}{Specifically, if the camera motion is with translation and fixed-axis rotation, 
we have the following certification theorem.}
\begin{theorem}[Robustness certification under camera motion with fixed-axis rotation]
\label{thm:resolvable_2d_proj}
Let $\alpha \in \mathcal{Z}\subset \mathbb{R}^6$ be  the parameters of projective transformation $\phi$ with translation $ (t_x, t_y, t_z)^T\in \mathbb{R}^3$ and fixed-axis rotation $ (\theta n_1, \theta n_2, \theta n_3)^T\in \mathbb{R}^3, \sum_{i=1}^3 n_i^2 = 1$,
suppose the composed camera motion $\varepsilon_1 \in \mathcal{Z}$  satisfies $ \phi(x, \varepsilon_1) =\phi(\phi(x, \varepsilon_0), \alpha) $ given some $\alpha \in \mathcal{Z}$ and zero-mean Gaussian motion $\varepsilon_0$ with variance $\sigma_x^2, \sigma_y^2, \sigma_z^2,\sigma_\theta^2$ for $t_x, t_y, t_z, \theta$ respectively,
        let $p_A,p_B\in[0,1]$ be bounds of the top-2 class probabilities for the motion smoothed model, i.e.,
            \begin{equation}
            \label{confidence_pab}
                q(y_A\mid x,\varepsilon_0) \geq p_A > p_B \geq \max_{y\neq y_A} q(y\mid x, \varepsilon_0).
                \end{equation}
        Then, it holds that $g(\phi(x, \alpha); \varepsilon_0) = g(x;\varepsilon_0)$ if $\alpha=(t_x, t_y, t_z,\theta n_1, \theta n_2, \theta n_3)^T$ satisfies
        \begin{equation}
        \label{equ:certify_condition}
            \sqrt{\left(\dfrac{\theta}{\sigma_\theta}\right)^2 + \left(\dfrac{t_x}{\sigma_x}\right)^2+\left(\dfrac{t_y}{\sigma_y}\right)^2+\left(\dfrac{t_z}{\sigma_z}\right)^2} < \dfrac{1}{2}\left(\Phi^{-1}(p_A) - \Phi^{-1}(p_B) \right).
        \end{equation}
\end{theorem}
\textcolor{revise}{The proof sketch is that based on Lemma \ref{lemma:resolvable_2d_proj}, the relative projection is resovable and compatible with the global projection, and specifically, $\gamma$ function in \eqref{lemma:gamma} will be additive $\gamma_{\alpha_1}(\alpha_2) = \alpha_1 + \alpha_2$ with fixed-axis rotation.   The full proof can be found in the appendix materials following previous certification work~\cite{cohen2019certified, li2021tss}. We remark that for the general rotation without a fixed axis, although $\varepsilon_1=\alpha + \varepsilon_0$ does not hold,  the $\gamma$ function can also be found based on multiplication in $SO(3)$ since Lemma \ref{lemma:resolvable_2d_proj} holds in general cases}. 

Since all the camera motions can be regarded as the composition of each 1-axis translation or rotation, we make the following corollary to show the certification for any 1-axis translation or rotation.
\begin{corollary}[Certification of 1-axis motion perturbation]
\label{coro:1-axis}
For any 1-axis camera motion perturbation $\alpha$ with non-zero entry $\alpha_i$ satisfying       
            $\alpha_i < \dfrac{\sigma_i}{2}\left(\Phi^{-1}(p_A) - \Phi^{-1}(p_B) \right)$ under motion smoothing $\varepsilon_0 \sim \mathcal{N}(0, \sigma_i^2)$, 
     it holds that $g(\phi(x, \alpha); \varepsilon_0) = g(x;\varepsilon_0) $ for the motion smoothed classifier $g$.
\end{corollary}
\begin{remark}
\label{remark}
Cor. \ref{coro:1-axis} directly follows Thm. \ref{thm:resolvable_2d_proj} by taking only one non-zero entry in $\alpha\in \mathcal{Z}\subset \mathbb{R}^6$, where each entry of in $\alpha=(t_x, t_y, t_z,\theta n_1, \theta n_2, \theta n_3)^T$ means $T_x$: {translation} along depth-orthogonal horizontal axis, $T_y$: {translation} along depth-orthogonal vertical axis, $T_z$: {translation} along depth axis, $R_x$: {rotation} around depth-orthogonal pitch axis, $R_y$: {rotation} around depth-orthogonal yaw axis and $R_z$: {rotation} around depth roll axis, as shown in Fig. \ref{fig:coordinates}.
\end{remark}
Therefore, we conduct comprehensive experiments for each 1-axis camera motion based on the Cor. \ref{coro:1-axis} and Remark \ref{remark}for robustness certification.

\section{Experiments}
In this section, we aim to address two questions: 1) How does the perturbation of camera motion influence the perception performance empirically? 2) How can we certify the accuracy using the motion-smoothed model under the camera perturbation within some radius? 
To answer these questions, we first set up a simulated indoor environment \textit{MetaRoom} with dense point cloud maps and conduct extensive experiments based on it. \textcolor{revise}{Besides, we also conduct real-world experiments on a robotic arm with an eye-in-hand camera to investigate  motion smoothing in robotic applications.}
Appendix materials contain more experiment details.

\subsection{Experimental Setup}

\begin{figure}
    \centering
    \includegraphics[width=0.13\textwidth]{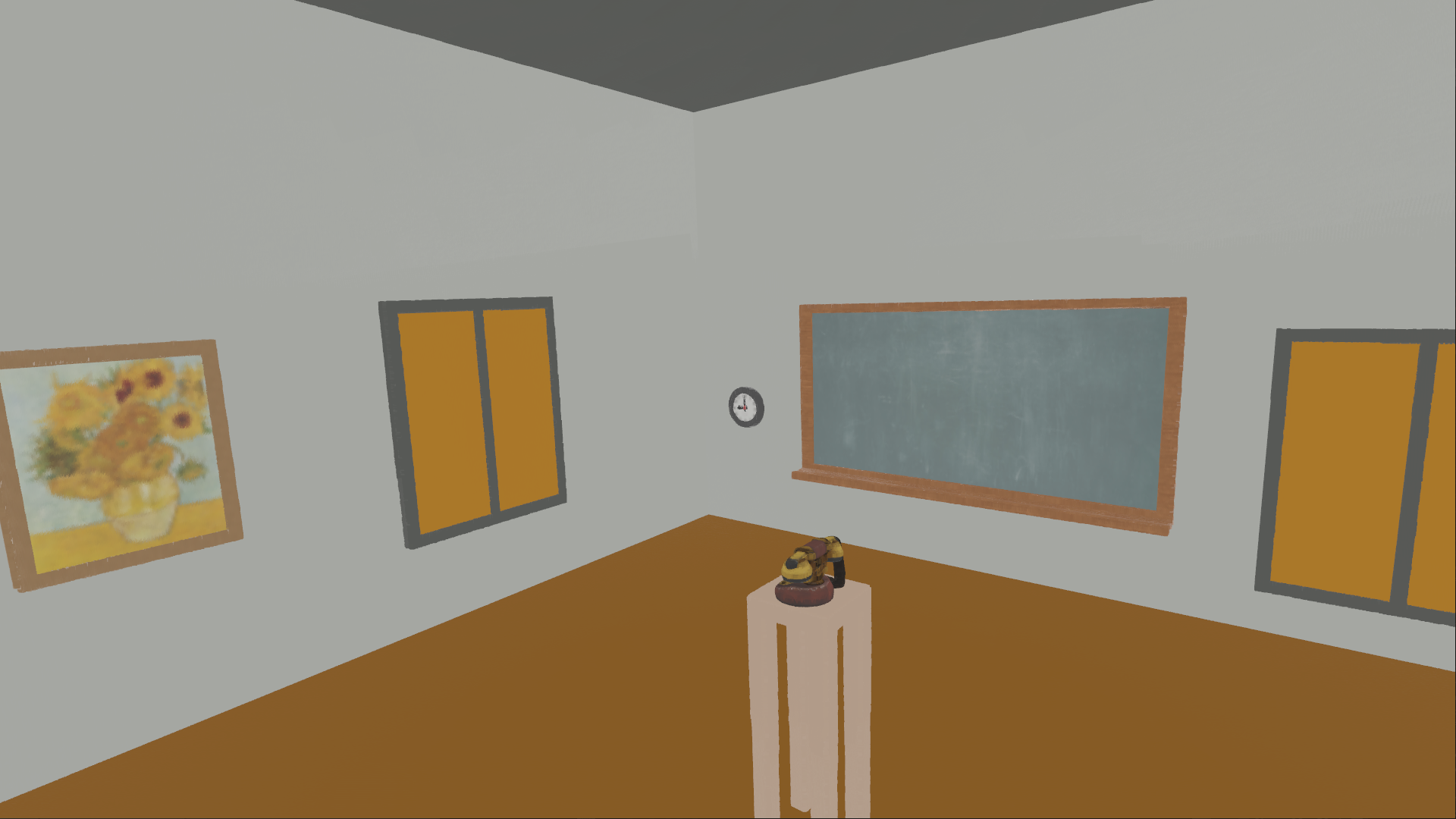}
    \includegraphics[width=0.13\textwidth]{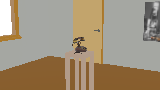}
    \includegraphics[width=0.13\textwidth]{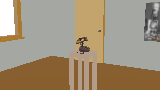}
    \includegraphics[width=0.13\textwidth]{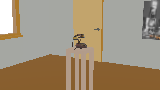}
    \includegraphics[width=0.13\textwidth]{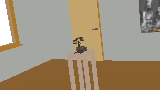}
    \includegraphics[width=0.13\textwidth]{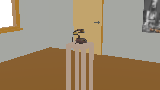}
    \includegraphics[width=0.13\textwidth]{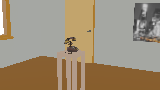} \\
    \includegraphics[width=0.13\textwidth]{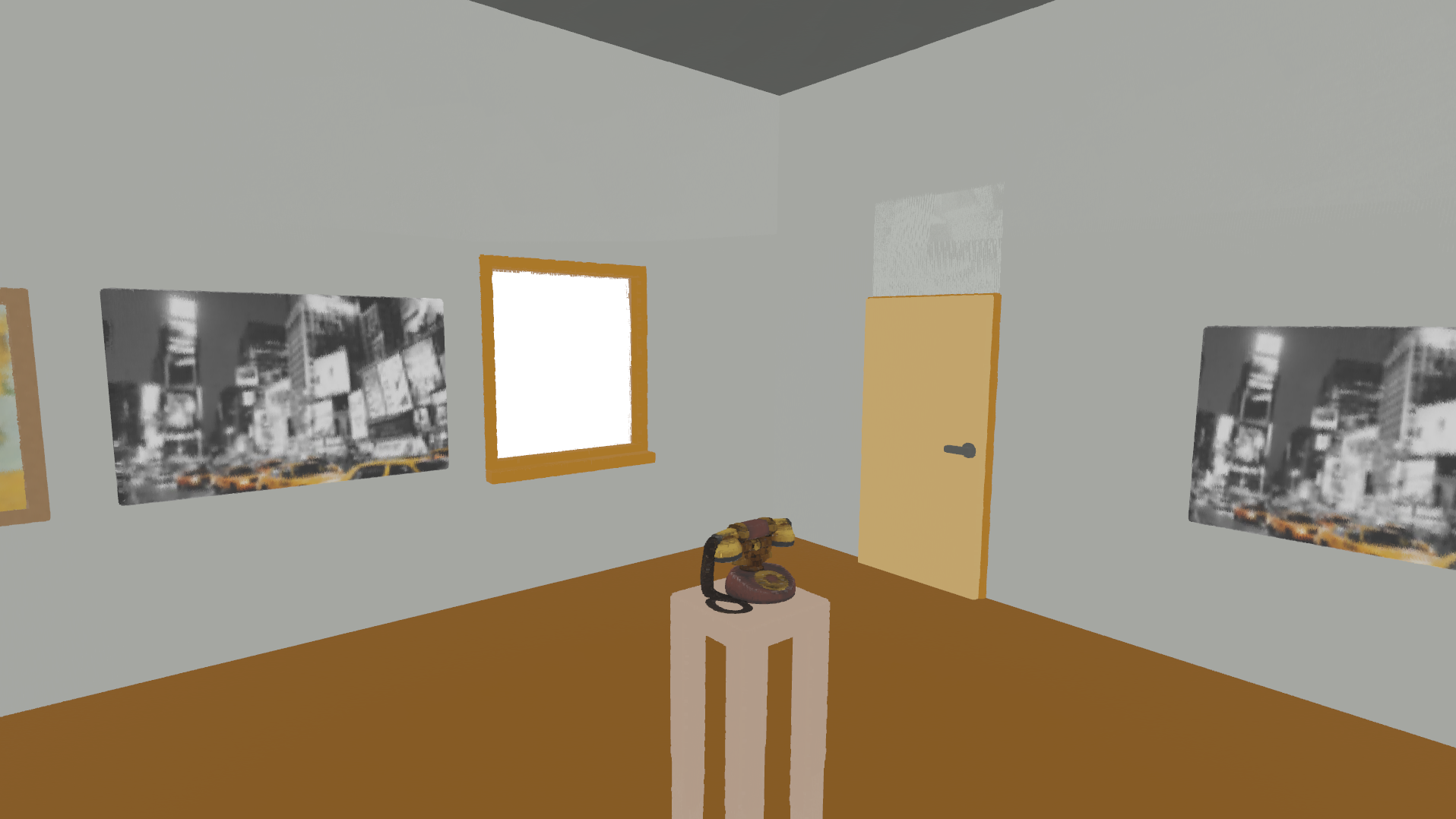}
    \includegraphics[width=0.13\textwidth]{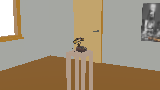}
    \includegraphics[width=0.13\textwidth]{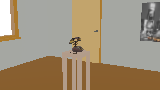}
    \includegraphics[width=0.13\textwidth]{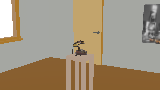}
    \includegraphics[width=0.13\textwidth]{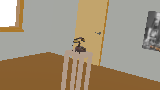}
    \includegraphics[width=0.13\textwidth]{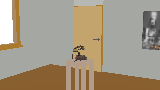}
    \includegraphics[width=0.13\textwidth]{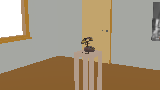} \\
    \vspace{2mm}
        \includegraphics[width=0.13\textwidth]{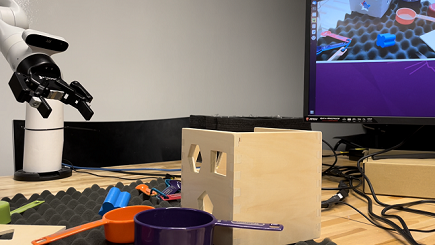}
    \includegraphics[width=0.13\textwidth]{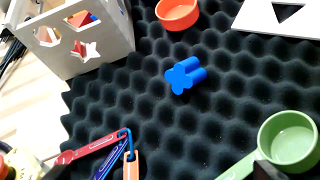}
    \includegraphics[width=0.13\textwidth]{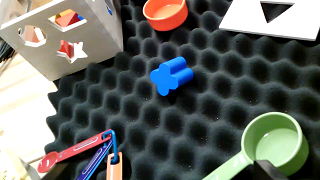}
    \includegraphics[width=0.13\textwidth]{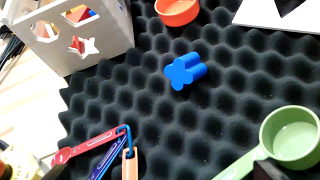}
    \includegraphics[width=0.13\textwidth]{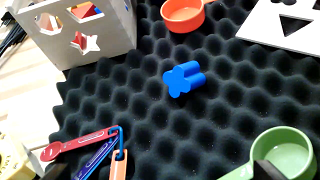}
    \includegraphics[width=0.13\textwidth]{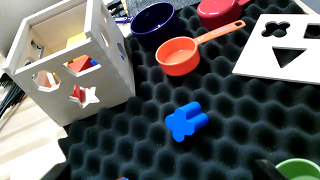}
    \includegraphics[width=0.13\textwidth]{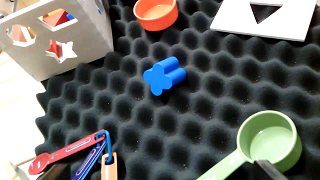} \\
    \includegraphics[width=0.13\textwidth]{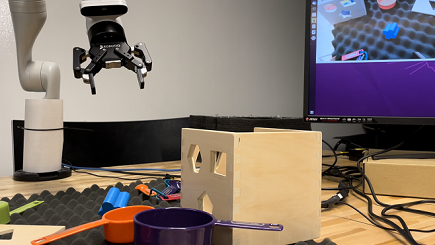}
    \includegraphics[width=0.13\textwidth]{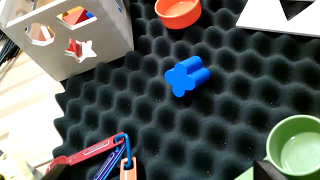}
    \includegraphics[width=0.13\textwidth]{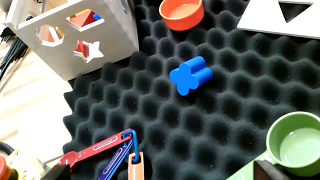}
    \includegraphics[width=0.13\textwidth]{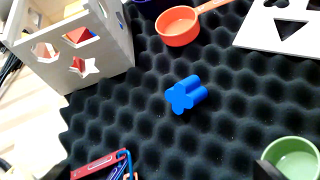}
    \includegraphics[width=0.13\textwidth]{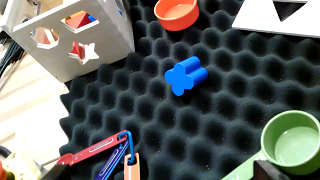}
    \includegraphics[width=0.13\textwidth]{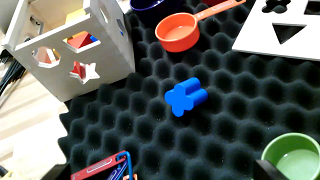}
    \includegraphics[width=0.13\textwidth]{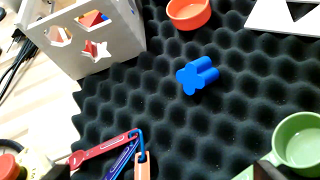} \\
    \vspace{-2mm}
    \caption{\textcolor{revise}{\small \textit{MetaRoom}  images under 6-axis perturbations (first two rows) and a real-world robot arm with an eye-in-hand camera (last two rows). From left to right: global scenarios, perturbations over $T_z$,  $T_x$, $T_y$,  $R_z$, $R_x$ and $R_y$.}}
    \label{fig:metaroom}
    \vspace{-5mm}
\end{figure}


\begin{wrapfigure}{r}{0.4\textwidth}
    \centering
    \includegraphics[width=0.39\textwidth]{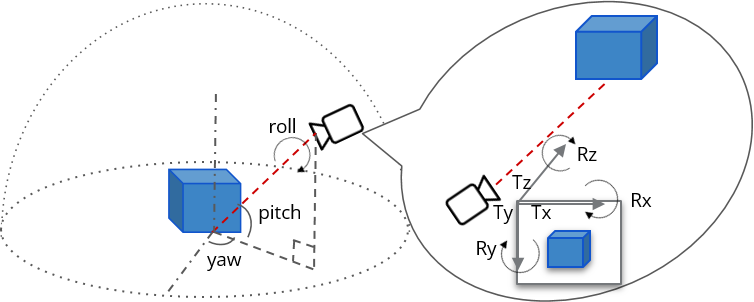}
    \caption{ \small The illustration of coordinates in \textcolor{revise}{\textit{MetaRoom} and real robot with camera motion.} }
    \vspace{-10pt}
    \label{fig:coordinates}
\end{wrapfigure}

\label{metaroom_setup}
\textbf{MetaRoom Dataset.} To implement the 3D-2D projection from point clouds to images through the camera, we first introduce the \textit{MetaRoom} dataset based on Webots~\cite{Webots}. The \textit{MetaRoom} dataset contains 20 commonly-seen indoor objects and for each object, we place it on a small table located in the center of an empty room with the size of $\SI{3}{m}\times \SI{3}{m}$. 
For each object, the collected data is associated with the point cloud of the entire room, all the camera poses, and camera intrinsic parameters. 
We first reconstruct the \SI{0.0025}{m}-density point clouds of the room together with each object on the table through random snapshots from an RGBD camera with 500 camera poses for training and 120 camera poses for testing. 
The objects are $\SI{0.1}{m}$ higher than the origin of the global coordinate. For the training set, the camera is oriented toward the origin and uniformly randomly moves within a semi-spherical area of $\SI{2.3}{m}\sim\SI{2.9}{m}$ radius with the yaw range of $360^\circ$, the pitch range of $0^\circ \sim 90^\circ$ and roll range of $-60^\circ \sim 60^\circ$. The poses in the test set are $\SI{2.8}{m}$ away from the original with 6 fixed yaw angles uniformly distributed in $0^\circ \sim 360^\circ$, the pitch angle of $10^\circ$, and  the roll angle of $0^\circ$. Note that the poses in the training set are at least $15^\circ$  away from any yaw, pitch, or roll angle of every pose in the test set to force the model to generalize rather than just memorizing the training data. The coordinate is illustrated in Fig. \ref{fig:coordinates}. 




\textbf{Model Training.} We adopt two off-the-shelf representative architectures, ResNet-18 and ResNet-50~\cite{he2016deep}, to train base classifiers on \textit{MetaRoom} for 100 epochs. 
For baselines,  we adopt motion-specific augmentation as a defense to train the \textbf{Motion Augmented} models while  \textbf{Vanilla} models are not trained with the augmentation since the augmented classifier could generalize well to the images with corresponding perturbation following \cite{li2021tss, chu2022tpc}. The motion augmentations are consistent with smoothing Gaussian distributions with $\sigma$ 
shown in Tab. \ref{tab:vanilla_certify}. Here motion augmented and vanilla are different strategies to train a \textbf{Base} classifier for 
\textbf{Smoothed} model  via motion smoothing in Def. \ref{def:smooth_parametric_classifier}. 

\begin{table}[]
\centering
\begin{tabular}{ccc}
\hline
Camera Motion Types &  Motion Aug. ResNet18 & Motion Aug. ResNet50  \\
\hline
$T_z$, radius {[}-0.1m, 0.1m{]} &  Base / Smoothed &  Base / Smoothed \\
\hline
Benign Accuracy                       &                \textbf{0.858}    /      0.842             &                     \textbf{0.875}    / 0.867                      \\
\textcolor{revise}{100-perturbed Emp. Robust Acc.}             &               \textcolor{revise}{0.758}     /        \textcolor{revise}{\textbf{0.817}}            &      \textcolor{revise}{0.775}    /      \textcolor{revise}{\textbf{0.825}}        \\
\hline
$T_x$, radius {[}-0.05m, 0.05m{]} &  Base / Smoothed &  Base / Smoothed\\
\hline
Benign Accuracy                       &        \textbf{0.925}           /    0.900              &                     \textbf{0.883}    /     0.867                   \\
\textcolor{revise}{100-perturbed Emp. Robust Acc.}               &         \textcolor{revise}{0.785}          /       \textcolor{revise}{\textbf{0.867}}            &       \textcolor{revise}{0.633}    /     \textcolor{revise}{\textbf{0.800}}       \\
\hline
$T_y$, radius {[}-0.05m, 0.05m{]} &  Base / Smoothed &  Base / Smoothed \\
\hline
Benign Accuracy                       &      \textbf{0.925}              /     0.892                &      0.917                  /     \textbf{0.942}                   \\
\textcolor{revise}{100-perturbed Emp. Robust Acc.}               &         \textcolor{revise}{0.808}           /          \textcolor{revise}{\textbf{0.842}}           &   \textcolor{revise}{0.842}        /  \textcolor{revise}{\textbf{0.908}}            \\
\hline
$R_z$, radius {[}-7$^\circ$, 7$^\circ${]} &  Base / Smoothed &  Base / Smoothed \\
\hline
Benign Accuracy                       &           0.933        /     \textbf{0.958}                 &        0.933               /     \textbf{0.950}                   \\
\textcolor{revise}{100-perturbed Emp. Robust Acc.}               &           \textcolor{revise}{0.875}         /           \textcolor{revise}{\textbf{0.892}}          &       \textcolor{revise}{0.867}    /   \textcolor{revise}{\textbf{0.917}}          \\
\hline
$R_x$, radius {[}-2.5$^\circ$, 2.5$^\circ${]} &  Base / Smoothed &  Base / Smoothed\\
\hline
Benign Accuracy                       &            \textbf{0.975}       /       0.950               &     0.925                  /      \textbf{0.942}                  \\
\textcolor{revise}{100-perturbed Emp. Robust Acc.}               &             \textcolor{revise}{\textbf{0.908} }      /          \textcolor{revise}{0.892}        &     \textcolor{revise}{0.850}        /       \textcolor{revise}{\textbf{0.917}}       \\
\hline
$R_y$, radius {[}-2.5$^\circ$, 2.5$^\circ${]} &  Base / Smoothed &  Base / Smoothed \\
\hline
Benign Accuracy                       &         0.917          /        \textbf{0.925}            &        0.975                /    \textbf{0.992}                    \\
\textcolor{revise}{100-perturbed Emp. Robust Acc.}               &           \textcolor{revise}{0.867}         /        \textcolor{revise}{\textbf{0.925}}           &        \textcolor{revise}{0.933}   /    \textcolor{revise}{\textbf{0.983}}         \\
\hline
\end{tabular}
\vspace{1mm}
\caption{\small  The comparison between  base and motion smoothed  models which are both trained with motion augmentations  in terms of benign and \textcolor{revise}{100-perturbed empirical robust  accuracy} for  all camera motions. The higher one between  each base and motion smoothed  model is in \textbf{bold}.}
\label{tab:base_smoothed}
    \vspace{-8mm}
\end{table}

\textbf{Evaluation Metrics.} 
The \textbf{Benign Accuracy} is calculated as the ratio of correctly classified projected images without any perturbation, which is used to show the robustness/accuracy trade-off~\cite{cohen2019certified}.
\textcolor{revise}{Based on the literature on spatial robustness \cite{engstrom2019exploring, sitawarin2022demystifying} that gradient-based attack methods \cite{goodfellow2014explaining, madry2017towards} do not perform better than grid-search-based attacks for spatial adversarial samples due to the highly non-convex optimization landscape in semantic transformation space, we  use grid search to evaluate the worst-case perturbations.
We uniformly sample 5 and 100 perturbed camera motions within the radius and consider the model is not robust if any of them
is wrongly classified, 
and report the ratio of robust ones over the whole test set as \textbf{5-perturbed Empirical Robust Accuracy} and \textbf{100-perturbed Empirical Robust Accuracy}, respectively. }
To provide a rigorous lower bound on the accuracy against any possible perturbations, we report \textbf{Certified Accuracy}~\cite{li2021tss, chu2022tpc} to evaluate the certification results, which is the fraction of test images that are both correctly classified and satisfy the certification condition of \eqref{equ:certify_condition} within motion perturbation radius through smoothing, meaning models will predict correctly with at least this certified accuracy for any camera perturbation within the given motion radius with high confidence. 
Following convention~\cite{cohen2019certified},  we use the confidence of 99\% and 1000 smoothing camera motions under zero-mean Gaussian distribution for each test image.

\begin{table}[ht]
\centering
\begin{tabular}{ccccc}
\hline
$\sigma_z$ &$T_z$  within radius {[}-0.1m, 0.1m{]} &  Benign Acc & \textcolor{revise}{5-pert Emp. Acc} & Certified  Acc\\
\hline
\multirow{4}{*}{0.1m}           & Smoothed Vanilla ResNet-18          &        \textbf{0.858}         &         0.817                   &   0.792 \\&
Smoothed Augmented ResNet-18               &        0.842        &         \textbf{0.833}                 &    \textbf{0.817}                     \\ \cline{2-5} &
Smoothed Vanilla ResNet-50                     &       0.675         &             0.617              &     0.558                      \\& 
Smoothed Augmented ResNet-50               &      \textbf{0.867}          &          \textbf{0.850}                &   \textbf{0.817} \\        \hline
$\sigma_x$  &$T_x$  within radius {[}-0.05m, 0.05m{]}&  Benign Acc & \textcolor{revise}{5-pert Emp. Acc} & Certified  Acc\\
\hline
\multirow{4}{*}{0.05m}              &  Smoothed Vanilla ResNet-18        &        0.825        &            0.783               &         0.700                  \\ &
Smoothed Augmented ResNet-18              &       \textbf{0.900}         &            \textbf{0.875}              &    \textbf{0.833}                      \\ \cline{2-5} &
Smoothed Vanilla ResNet-50                     &     0.767            &        0.675                   &        0.508                   \\&
Smoothed Augmented ResNet-50                &      \textbf{0.867}           &          \textbf{0.825}                &   \textbf{0.708} \\        \hline
$\sigma_y$ & $T_y$  within radius {[}-0.05m, 0.05m{]} & Benign Acc & \textcolor{revise}{5-pert Emp. Acc} & Certified  Acc\\
\hline
\multirow{4}{*}{0.05m}                & Smoothed Vanilla ResNet-18      &        0.850         &            0.825              &         0.767                  \\ &
Smoothed Augmented ResNet-18    &                \textbf{0.892}           &      \textbf{0.875}                     &        \textbf{0.817}                  \\ \cline{2-5} &
Smoothed Vanilla ResNet-50                    &       0.792          &        0.767                   &           0.683               \\& 
Smoothed Augmented ResNet-50               &        \textbf{0.942}         &            \textbf{0.925}              &  \textbf{0.892}  \\        \hline
$\sigma_\theta$ &$R_z$  within radius {[}-7$^\circ$, 7$^\circ${]}&  Benign Acc & \textcolor{revise}{5-pert Emp. Acc} & Certified  Acc\\
\hline
\multirow{4}{*}{7$^\circ$}                & Smoothed Vanilla ResNet-18      &       0.817          &              0.742             &      0.608                     \\ &
Smoothed Augmented ResNet-18   &               \textbf{0.958}            &              \textbf{0.933}             &        \textbf{0.883}                 \\ \cline{2-5} &
Smoothed Vanilla ResNet-50                     &        0.758         &          0.717                 &     0.633                      \\ &
Smoothed Augmented ResNet-50               &       \textbf{0.950}          &               \textbf{0.917}&  \textbf{0.883}  \\        \hline
$\sigma_\theta$ &$R_x$   within radius {[}-2.5$^\circ$, 2.5$^\circ${]} & Benign Acc & \textcolor{revise}{5-pert Emp. Acc} & Certified  Acc\\
\hline
 \multirow{4}{*}{2.5$^\circ$}              &  Smoothed Vanilla ResNet-18        &      0.842         &                0.800           &       0.633                    \\&
Smoothed Augmented ResNet-18               &         \textbf{0.950}       &                  \textbf{0.942}        &            \textbf{0.867}              \\ \cline{2-5} &
Smoothed Vanilla ResNet-50                     &       0.767          &         0.742                  &       0..567                    \\& 
Smoothed Augmented ResNet-50               &        \textbf{0.942}         &          \textbf{0.933}                &   \textbf{0.867} \\        \hline
$\sigma_\theta$ &$R_y$   within radius {[}-2.5$^\circ$, 2.5$^\circ${]}&  Benign Acc & \textcolor{revise}{5-pert Emp. Acc} & Certified  Acc\\
\hline
\multirow{4}{*}{2.5$^\circ$}              &   Smoothed Vanilla ResNet-18        &    0.892           &          0.875                 &         0.708                 \\&
Smoothed Augmented ResNet-18               &      \textbf{0.925}           &                \textbf{0.925}          &          \textbf{0.917}               \\ \cline{2-5} &
Smoothed Vanilla ResNet-50                     &       0.808          &       0.783                    &        0.717                  \\& 
Smoothed Augmented ResNet-50      &            \textbf{0.992}              &        \textbf{0.992}                   & \textbf{0.967}  \\        \hline
\end{tabular}
\vspace{1mm}
\caption{\small The comparison between smoothed vanilla and smoothed augmented models in terms of benign, \textcolor{revise}{5-perturbed empirical robust (5-perb Emp.)}, and certified  accuracy for  with all camera motions. The higher one between  vanilla and motion augmented models is in \textbf{bold}.}
\label{tab:vanilla_certify}
\vspace{-6mm}
\end{table}

\subsection{Empirical Robustness against Camera Motion with Smoothing}
We answer the question about the influence of camera motion perturbation on the perception performance empirically. For models trained with motion augmented defense, we compare the  base and smoothed classifiers over benign and empirical robust accuracy  in Tab. \ref{tab:base_smoothed}. 
It can be seen that under each motion perturbation, the empirical robust accuracy is lower than the benign accuracy for the base and smoothed models, showing that the perception models are not robust under motion perturbations even with the motion augmented defense when training. 
However, the gap between empirical robust accuracy and  benign accuracy of motion smoothed models become less  than those of base models due to the effective motion smoothing that improves robustness.
Specifically, the motion smoothed models perform better in terms of empirical robust accuracy than the base models for camera motion perturbation along all axis. Interestingly, under a similar magnitude of radius, rotational smoothing will mostly increase the benign accuracy while translational smoothing tends to decrease the benign accuracy due to the robustness/accuracy trade-off  \cite{cohen2019certified}. \textcolor{revise}{The comparison and analysis between 5-perturbed and 100-perturbed empirical robust accuracy can be found in Appendix Section \ref{details} and Table \ref{100_perturb}.}

\subsection{Comparison of Certified Accuracy}
Here we aim to show how we certify the robust accuracy using the motion smoothing strategy against motion perturbation, so we compare the smoothed augmented models with the smoothed vanilla models in Tab. \ref{tab:vanilla_certify}. For the motion perturbations on each axis, under the same smoothing strategy, most smoothed augmented models have better performance in benign, empirical robust and certified accuracy  compared to smoothed vanilla models, showing that the motion augmentation improves the overall perception performance, empirical and certifiable robustness. 
Besides, the certified accuracy is only slightly lower than empirical robust accuracy under the same perturbation radius, which shows that our certification guarantees are effective and scalable against all camera motion perturbations for different baseline models.

\subsection{Ablation Study}

\begin{figure}
    \centering
    \includegraphics[width=0.3\textwidth]{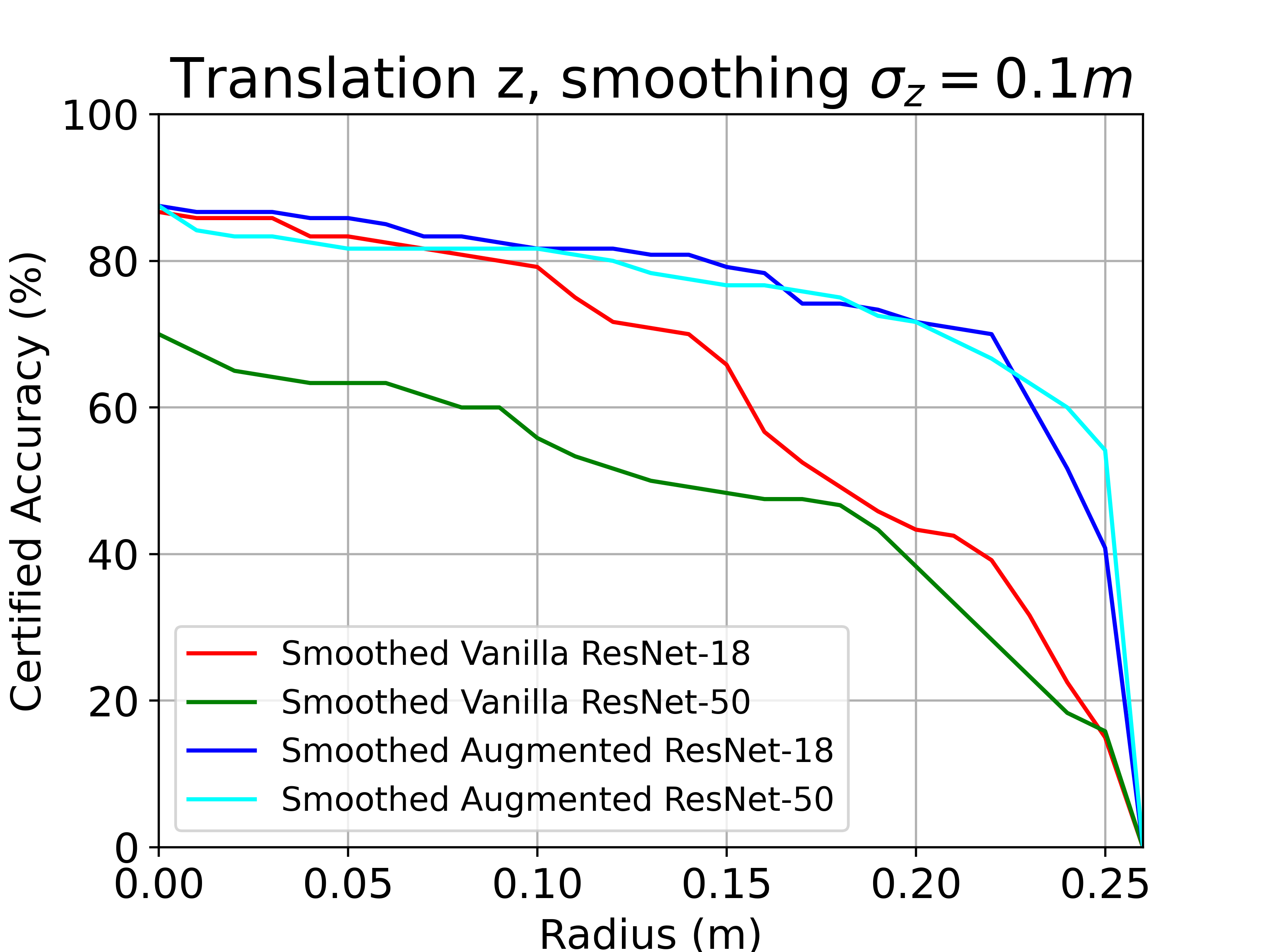}
    \includegraphics[width=0.3\textwidth]{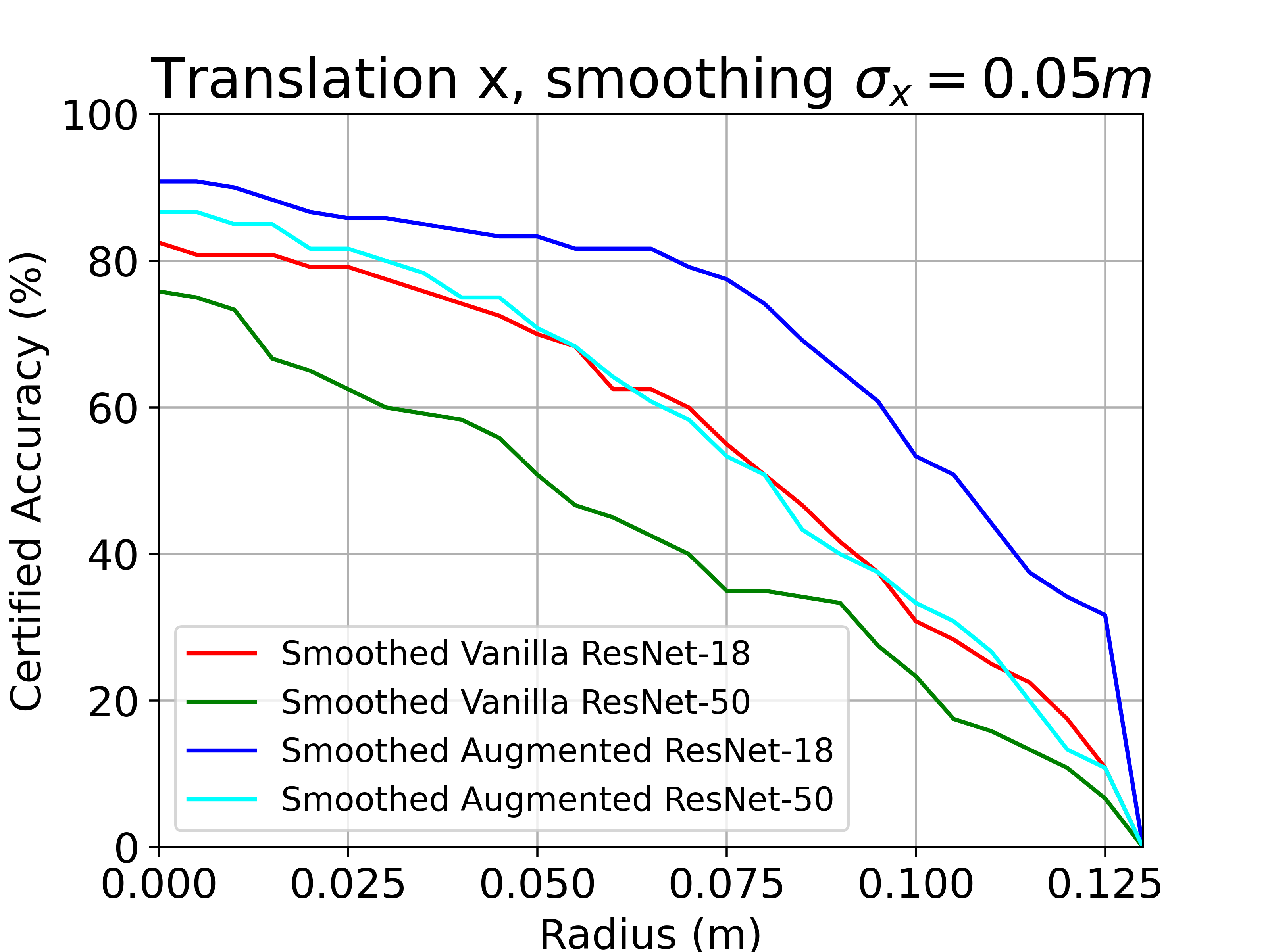}
    \includegraphics[width=0.3\textwidth]{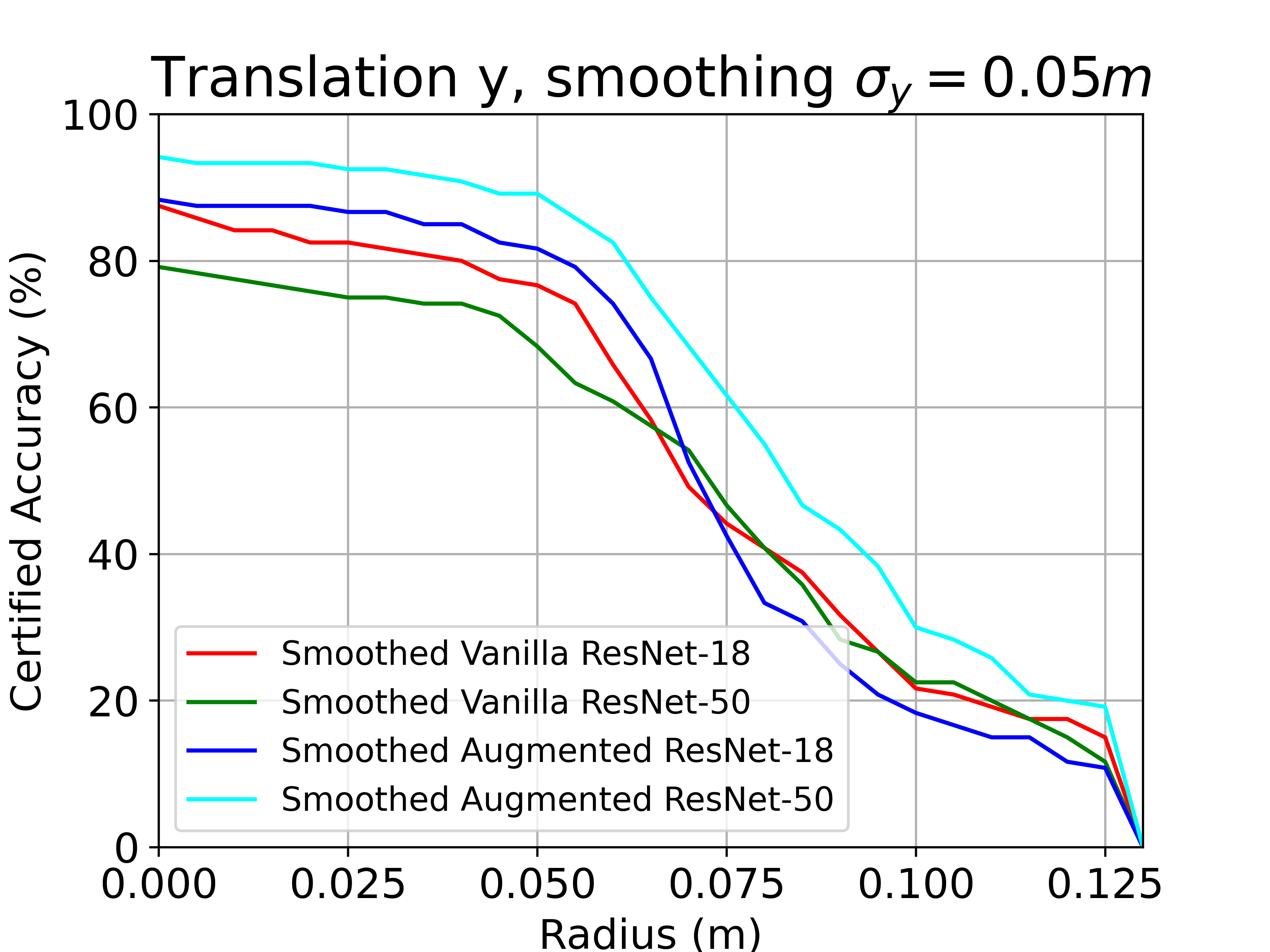}\\
    \includegraphics[width=0.3\textwidth]{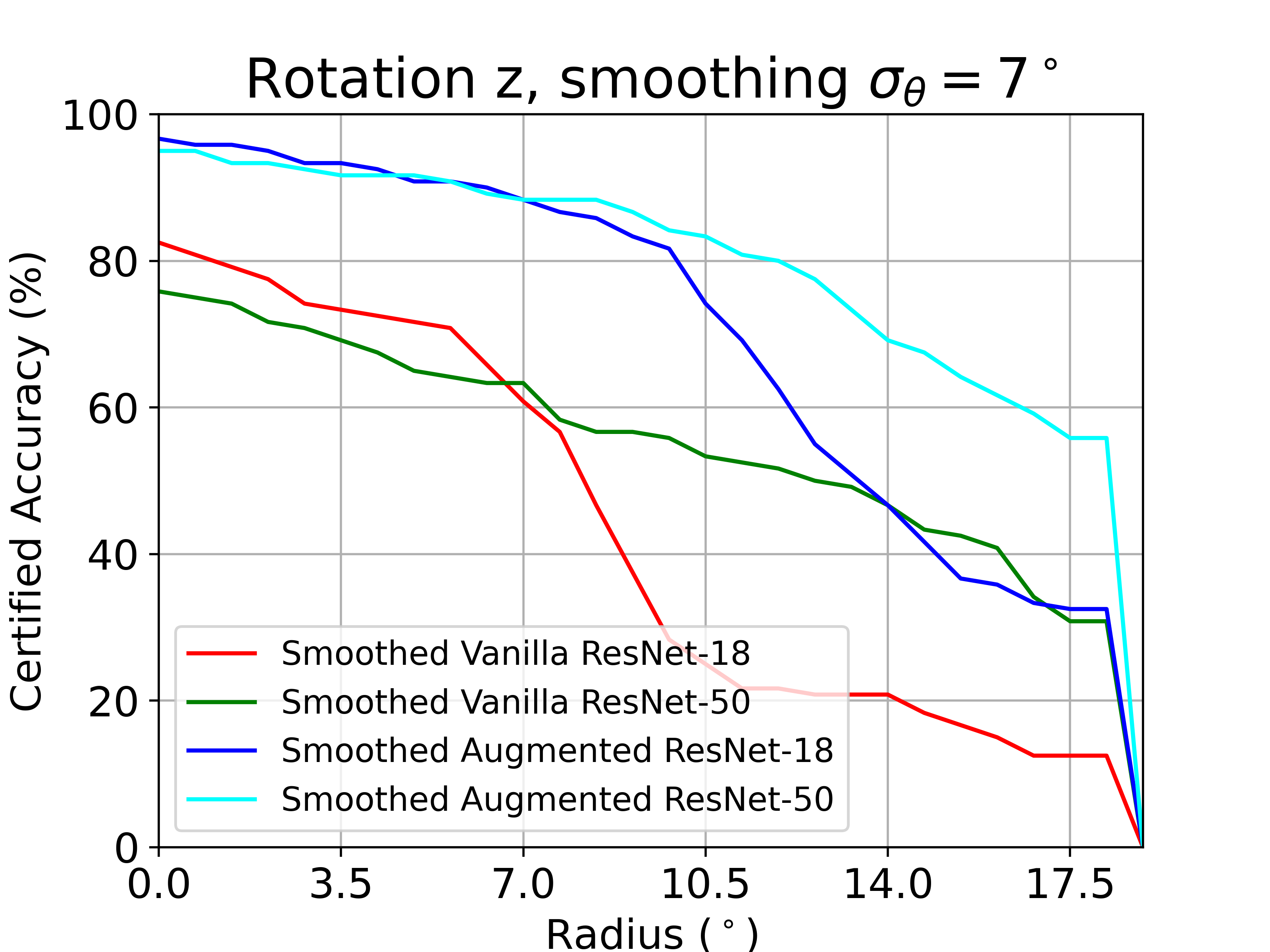}
    \includegraphics[width=0.3\textwidth]{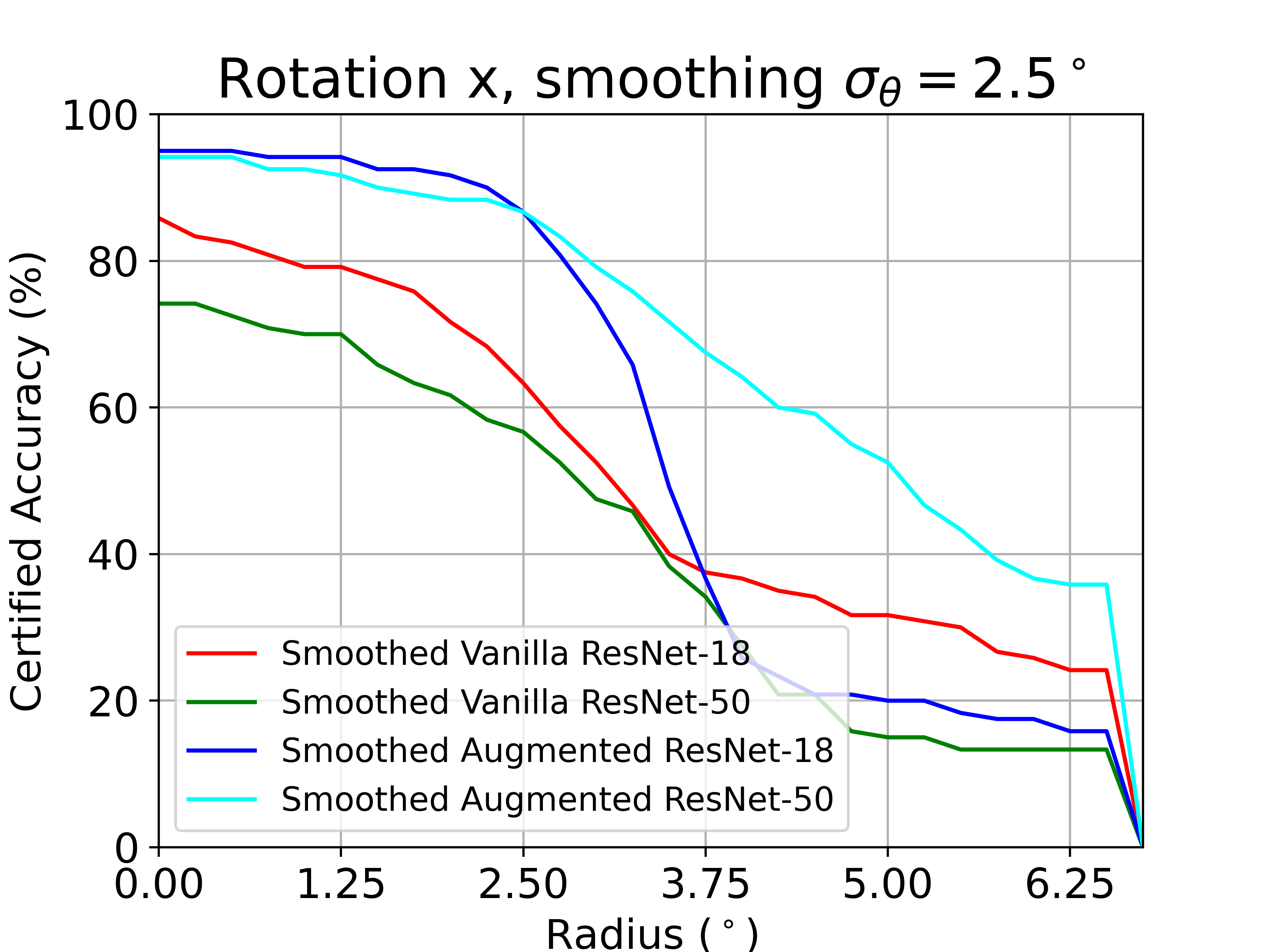}
    \includegraphics[width=0.3\textwidth]{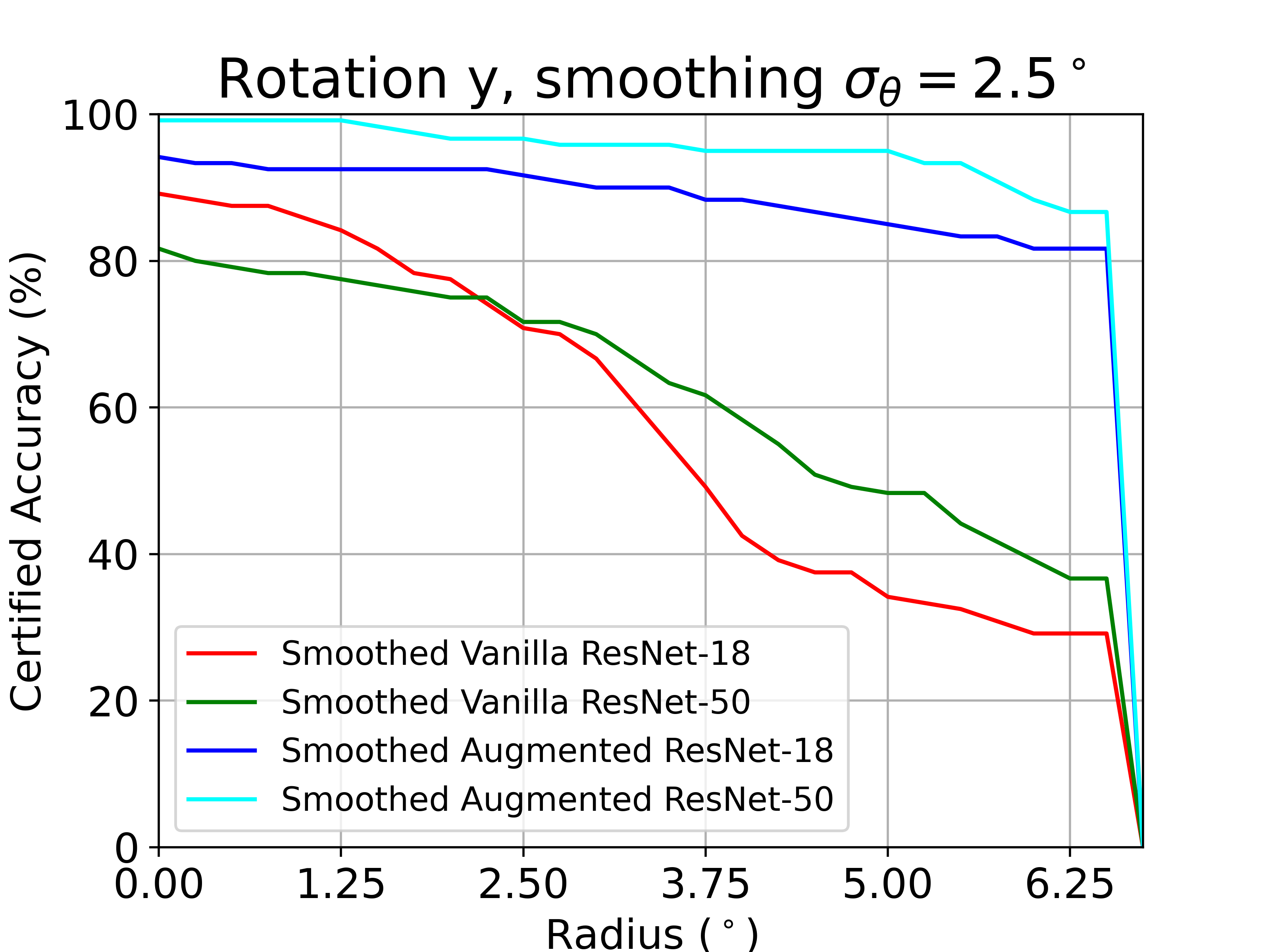}
    \vspace{-2mm}
    \caption{\small Certified accuracy with respect to radius for all ego-motions}
    \label{fig:radius}
\end{figure}

\textbf{Certified accuracy under different radius.} Fig. \ref{fig:radius} shows the certified accuracy with respect to the motion perturbation radius on each axis. We find that as the certified radius increases, the certified accuracy decreases and comes to 0 at some radius. Smoothed vanilla models have slowly decaying  certified accuracy. For motion augmented models,  the certified accuracies beyond the certified radii of \textit{Translation z} and \textit{Rotation y} still remain high until two times larger than $\sigma$ in motion smoothing, while \textit{Translation y} and \textit{Rotation x} have quickly decayed certified accuracies, showing that \textit{Translation z} and \textit{Rotation y} can be better certified over larger perturbation radii. 

\textbf{Comparison of different model complexity.} Smoothed vanilla ResNet-18  performs better in certified accuracy compared to ResNet-50 within smoothing radii $\sigma$. With motion augmented defense, although most empirical robust accuracy of base ResNet-50 is worse than base ResNet-18 in Tab. \ref{tab:base_smoothed} and \ref{tab:base_smoothed_vanilla}, ResNet-50 has better certified accuracy after motion smoothing under larger perturbation radii for most 1-axis camera motions in Fig. \ref{fig:radius}, which indicates that motion-augmented smoothed classifiers with larger complexity tend to be more certifiably robust to camera motion perturbations, although they may suffer from lower empirical robust accuracy due to overfitting. 

\begin{table}[]
\centering
\begin{tabular}{cccccccc}
\hline
\multirow{2}{*}{\textcolor{revise}{Model}} & \multirow{2}{*}{\textcolor{revise}{Benign Accuracy}} & \multicolumn{6}{c}{\textcolor{revise}{10-perturbed Empirical Robust Accuracy}  }                                                                                  \\ \cline{3-8} 
                       &                                  & \textcolor{revise}{$ T_x$}  & \textcolor{revise}{$ T_y$}  & \textcolor{revise}{$ T_z$}  & \textcolor{revise}{$ R_x$}  & \textcolor{revise}{$ R_y$}  & \textcolor{revise}{$ R_z$}  \\ \hline
\textcolor{revise}{Base}                   & \multirow{2}{*}{\textcolor{revise}{83.3\%}}          &\textcolor{revise}{ 76.3\%}                   & \textcolor{revise}{78.1\%    }               & \textcolor{revise}{79.8\%}                   & \textcolor{revise}{77.1\%}             &\textcolor{revise}{ 74.6\%}             & \textcolor{revise}{82.5\% }            \\
\textcolor{revise}{Smoothed}               &                                  & \textcolor{revise}{\textbf{78.9\%  }}                 & \textcolor{revise}{\textbf{81.6\% }}                  & \textcolor{revise}{\textbf{82.5\%}}                   & \textcolor{revise}{\textbf{77.2\% }}            & \textcolor{revise}{\textbf{75.4\%} }            & \textcolor{revise}{\textbf{83.3\%}}             \\ \hline
\end{tabular}
\vspace{1mm}
\caption{\small \textcolor{revise}{Quantitative results of real-world robotic perception model. The perturbation range of translations ($T_x, T_y, T_z$) is $[-1.25cm, 1.25cm]$ and perturbation range of rotations ($R_x, R_y, R_z$) is $[-2.5^\circ, 2.5^\circ]$. The variance of zero-mean Gaussian smoothing distribution is $0.625cm$ translations and $1.25^\circ$ for rotations. The higher values between the base and smoothed ones are in \textbf{bold}.}}
\label{tab:real_robot}
\vspace{-6mm}
\end{table}

\textcolor{revise}{\subsection{Real-world Experiments}}
\textcolor{revise}{\textbf{Experiment setup.} We conduct hardware robotic experiments using the Kinova-Gen3 Arm of 7-DoF with an  eye-in-hand camera in the pick-place task environment. These 6 objects with different shapes and colors and the perception model deployed on the arm is based on the ResNet50 classification model. For the data collection, following the spherical coordinate in Figure \ref{fig:coordinates}, we first place each object $0.7m$ away from the robot base and randomly choose roll angles in $[-60^\circ, 60^\circ]$, pitch angles in $[35^\circ, 65^\circ]$, yaw angles in $[-30^\circ, 30^\circ]$ and radius in $[0.35m, 0.45m]$, capturing 2500 images along all the random waypoints using the default planning trajectories for each object. 
The non-overlapped gap is set between the training set  and test set to choose 19 random poses, which are fixed for 6 objects as 114 test poses in total.
 The base model has well-trained over 5 epochs. At each perturbation, the smoothed model is with the 10 samples from zero-mean Gaussian distribution with variance $0.625cm$ for all translations and $1.25^\circ$ for all rotations. Following the metrics in Section \ref{metaroom_setup}, for both the base model and smoothed model, we adopt 10 uniform samples over  $[-1.25cm, 1.25cm]$ and  $[-2.5^\circ, 2.5^\circ]$ as empirical robust accuracy. We report benign accuracy without any perturbation using the base model for comparison and omit certification to evaluate empirical robustness in real robot applications. More details can be found in Appendix Section \ref{sec:real_robot} }

\begin{figure}
    \centering
    \includegraphics[width=0.19\textwidth]{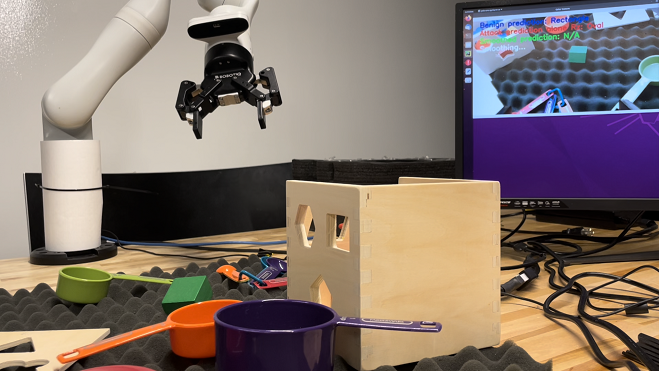}
    \includegraphics[width=0.19\textwidth]{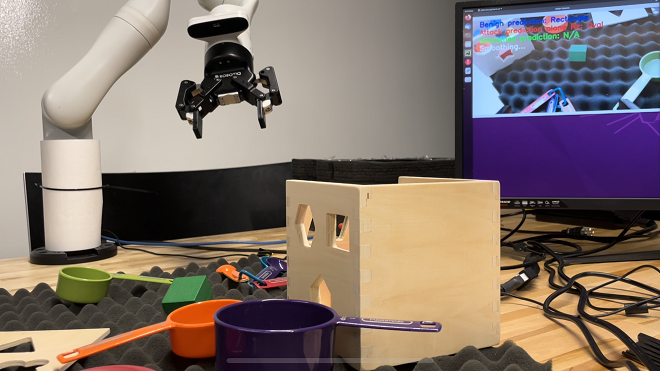}
    \includegraphics[width=0.19\textwidth]{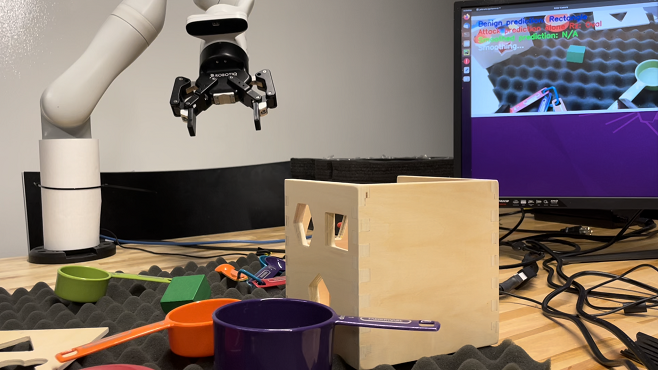}
    \includegraphics[width=0.19\textwidth]{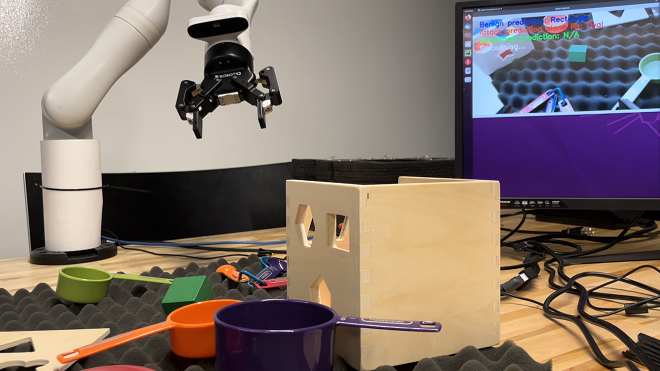}
    \includegraphics[width=0.19\textwidth]{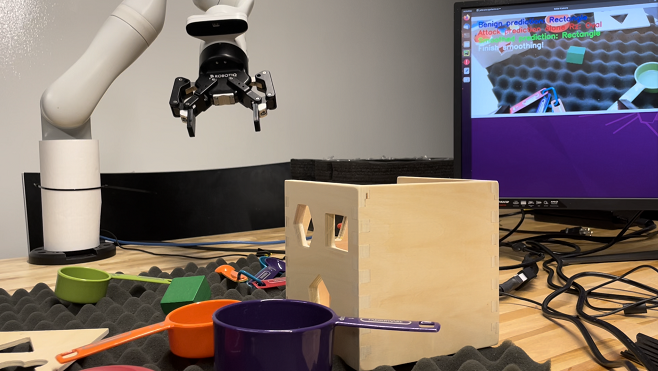}\\
    \includegraphics[width=0.19\textwidth]{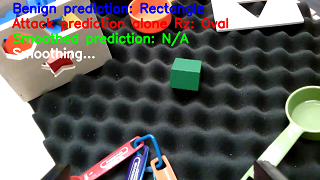}
    \includegraphics[width=0.19\textwidth]{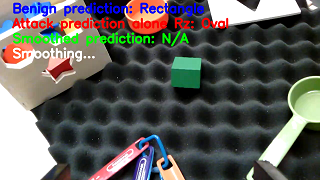}
    \includegraphics[width=0.19\textwidth]{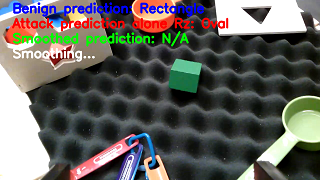}
    \includegraphics[width=0.19\textwidth]{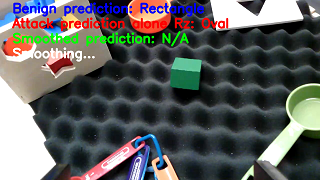}
    \includegraphics[width=0.19\textwidth]{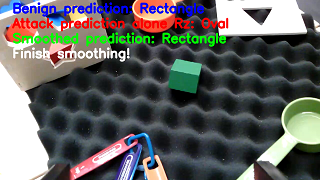}\\ 
    \vspace{2mm}
    \includegraphics[width=0.19\textwidth]{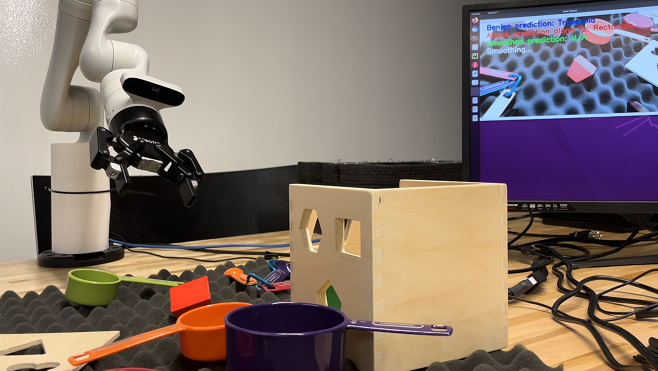}
    \includegraphics[width=0.19\textwidth]{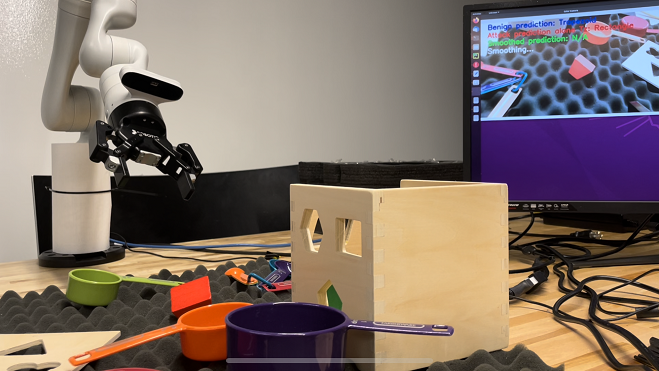}
    \includegraphics[width=0.19\textwidth]{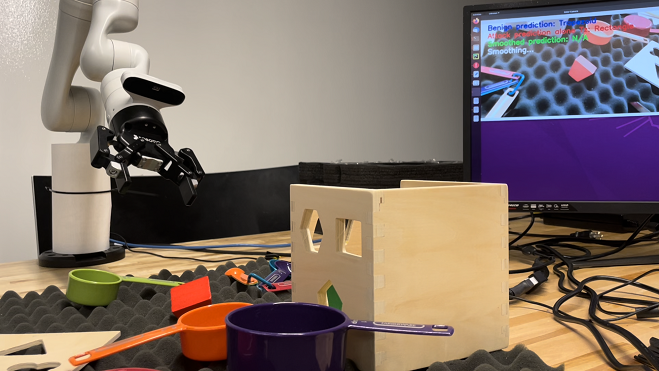}
    \includegraphics[width=0.19\textwidth]{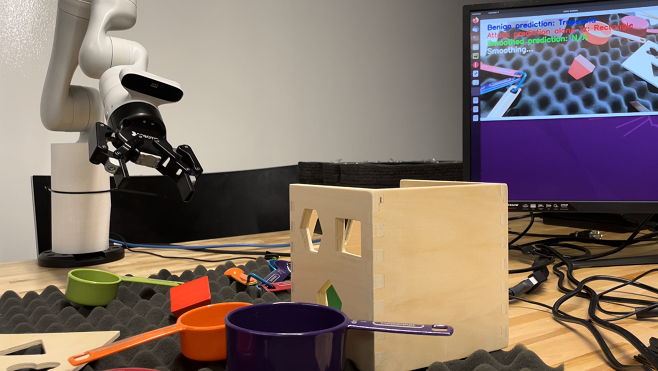}
    \includegraphics[width=0.19\textwidth]{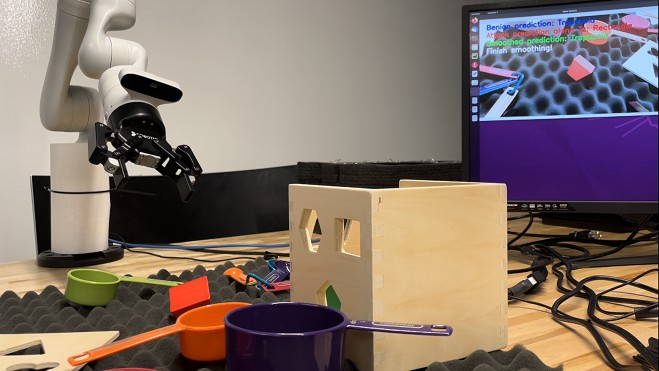}\\
    \includegraphics[width=0.19\textwidth]{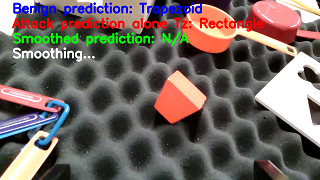}
    \includegraphics[width=0.19\textwidth]{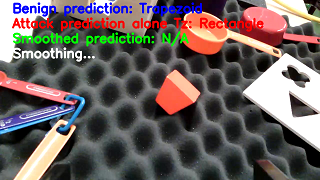}
    \includegraphics[width=0.19\textwidth]{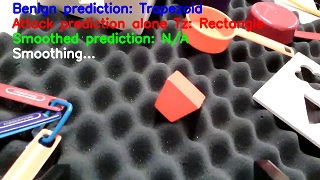}
    \includegraphics[width=0.19\textwidth]{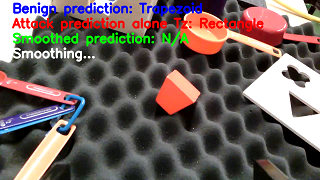}
    \includegraphics[width=0.19\textwidth]{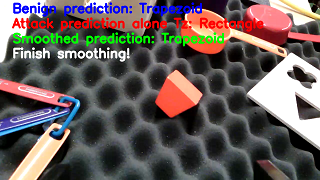}\\
    \vspace{-2mm}
    \caption{\textcolor{revise}{\small Smoothing process to improve robustness against camera motion of $R_z$ (top) and $T_z$ (bottom). The left four columns are randomized smoothing samples, and the right column is the classification result after smoothing.}}
    \label{fig:smoothing}
    \vspace{-2mm}
\end{figure}

\textcolor{revise}{\textbf{Results and analysis.} From Table \ref{tab:real_robot}, it can be seen that the 10-perturbed empirical robust accuracy is lower than the benign accuracy for the base model, which means the small perturbations along each axis do influence the performance of the real-world robotic perception. Besides, our proposed smoothed model improves the robust accuracy against all the perturbations, and results of $T_z$ and $R_z$ after smoothing are very close to benign accuracy.  Since the perturbations are small, the perturbation influence is not that significant. But the smoothing variance is also  tiny enough to be reasonable and practical in robotic applications, the improvement is observable and eligible to validate the effectiveness of the smoothing method. Besides, we present the qualitative results in Figure \ref{fig:smoothing} to illustrate the smoothing process to improve the perception robustness.} 


\subsection{Limitation and Discussion}
One limitation of this work is that our current certification framework is only evaluated on image classification tasks, although it is fundamental for other robot applications. It can be extended to regression tasks through discretization and applied to object detection, keypoint detection, depth estimation, etc.
In addition, the robustness certification procedure requires the prior 3D dense point cloud of the entire environment, which may be hard to obtain sometimes.
It also costs many computational resources to obtain the guarantee  and can be addressed through differential certifications \cite{li2020sok, li2021tss} as future work.
Finally, the current certification framework is built upon the camera sensor, and it would be interesting to extend the current work to other perception sensors in robotics and autonomous driving, e.g. 3D LiDAR.


\section{Conclusion}
 In this work, we study the robust visual perception against camera motion perturbation as the projective transformation from 3D to 2D. We propose a robustness certification framework via a camera motion smoothing approach to provide robust guarantees for image classification models for real-world robotic applications.  
 We collect a realistic indoor robotic dataset MetaRoom with the dense point cloud map for robustness certification against camera motion perturbation.  We conduct extensive experiments to compare the empirical robust accuracy with the certified robust accuracy for the motion smoothed model within  large radii of 6-axis camera motion perturbations, i.e., translations and rotations, to guarantee the lower bounds of accuracy  for trustworthy robotic perception. \textcolor{revise}{We further conduct real-world robot experiments to show the improvement of the smoothing method in robotic applications.}

\clearpage
\acknowledgments{We gratefully acknowledge support from the National Science Foundation under grant CAREER CNS-2047454. We also would like to thank Prof. Bo Li from UIUC for the thoughtful feedback and Shiqi Liu from CMU for helping to conduct the real robot experiment.}


\bibliography{ref}  

\begin{thebibliography}{54}
\providecommand{\natexlab}[1]{#1}
\providecommand{\url}[1]{\texttt{#1}}
\expandafter\ifx\csname urlstyle\endcsname\relax
  \providecommand{\doi}[1]{doi: #1}\else
  \providecommand{\doi}{doi: \begingroup \urlstyle{rm}\Url}\fi

\bibitem[Iandola et~al.(2016)Iandola, Han, Moskewicz, Ashraf, Dally, and
  Keutzer]{iandola2016squeezenet}
F.~N. Iandola, S.~Han, M.~W. Moskewicz, K.~Ashraf, W.~J. Dally, and K.~Keutzer.
\newblock Squeezenet: Alexnet-level accuracy with 50x fewer parameters and< 0.5
  mb model size.
\newblock \emph{arXiv preprint arXiv:1602.07360}, 2016.

\bibitem[Hu et~al.(2020)Hu, Yang, Qiao, Zhao, and Wang]{hu2020seasondepth}
H.~Hu, B.~Yang, Z.~Qiao, D.~Zhao, and H.~Wang.
\newblock Seasondepth: Cross-season monocular depth prediction dataset and
  benchmark under multiple environments.
\newblock \emph{arXiv preprint arXiv:2011.04408}, 2020.

\bibitem[Deng et~al.(2009)Deng, Dong, Socher, Li, Li, and
  Fei-Fei]{deng2009imagenet}
J.~Deng, W.~Dong, R.~Socher, L.-J. Li, K.~Li, and L.~Fei-Fei.
\newblock Imagenet: A large-scale hierarchical image database.
\newblock In \emph{2009 IEEE conference on computer vision and pattern
  recognition}, pages 248--255. Ieee, 2009.

\bibitem[He et~al.(2016)He, Zhang, Ren, and Sun]{he2016deep}
K.~He, X.~Zhang, S.~Ren, and J.~Sun.
\newblock Deep residual learning for image recognition.
\newblock In \emph{Proceedings of the IEEE conference on computer vision and
  pattern recognition}, pages 770--778, 2016.

\bibitem[Dosovitskiy et~al.(2020)Dosovitskiy, Beyer, Kolesnikov, Weissenborn,
  Zhai, Unterthiner, Dehghani, Minderer, Heigold, Gelly,
  et~al.]{dosovitskiy2020image}
A.~Dosovitskiy, L.~Beyer, A.~Kolesnikov, D.~Weissenborn, X.~Zhai,
  T.~Unterthiner, M.~Dehghani, M.~Minderer, G.~Heigold, S.~Gelly, et~al.
\newblock An image is worth 16x16 words: Transformers for image recognition at
  scale.
\newblock \emph{arXiv preprint arXiv:2010.11929}, 2020.

\bibitem[Redmon et~al.(2016)Redmon, Divvala, Girshick, and
  Farhadi]{redmon2016you}
J.~Redmon, S.~Divvala, R.~Girshick, and A.~Farhadi.
\newblock You only look once: Unified, real-time object detection.
\newblock In \emph{Proceedings of the IEEE conference on computer vision and
  pattern recognition}, pages 779--788, 2016.

\bibitem[Xu et~al.(2022{\natexlab{a}})Xu, Xiang, Xia, Han, Li, and
  Ma]{xu2022opv2v}
R.~Xu, H.~Xiang, X.~Xia, X.~Han, J.~Li, and J.~Ma.
\newblock Opv2v: An open benchmark dataset and fusion pipeline for perception
  with vehicle-to-vehicle communication.
\newblock In \emph{2022 International Conference on Robotics and Automation
  (ICRA)}, pages 2583--2589. IEEE, 2022{\natexlab{a}}.

\bibitem[Xu et~al.(2022{\natexlab{b}})Xu, Xiang, Tu, Xia, Yang, and
  Ma]{xu2022v2x}
R.~Xu, H.~Xiang, Z.~Tu, X.~Xia, M.-H. Yang, and J.~Ma.
\newblock V2x-vit: Vehicle-to-everything cooperative perception with vision
  transformer.
\newblock \emph{arXiv preprint arXiv:2203.10638}, 2022{\natexlab{b}}.

\bibitem[He et~al.(2017)He, Gkioxari, Doll{\'a}r, and Girshick]{he2017mask}
K.~He, G.~Gkioxari, P.~Doll{\'a}r, and R.~Girshick.
\newblock Mask r-cnn.
\newblock In \emph{Proceedings of the IEEE international conference on computer
  vision}, pages 2961--2969, 2017.

\bibitem[Hu et~al.(2020)Hu, Qiao, Cheng, Liu, and Wang]{hu2020dasgil}
H.~Hu, Z.~Qiao, M.~Cheng, Z.~Liu, and H.~Wang.
\newblock Dasgil: Domain adaptation for semantic and geometric-aware
  image-based localization.
\newblock \emph{IEEE Transactions on Image Processing}, 30:\penalty0
  1342--1353, 2020.

\bibitem[Xu et~al.(2022)Xu, Tu, Xiang, Shao, Zhou, and Ma]{xu2022cobevt}
R.~Xu, Z.~Tu, H.~Xiang, W.~Shao, B.~Zhou, and J.~Ma.
\newblock Cobevt: Cooperative bird's eye view semantic segmentation with sparse
  transformers.
\newblock \emph{arXiv preprint arXiv:2207.02202}, 2022.

\bibitem[Eykholt et~al.(2018)Eykholt, Evtimov, Fernandes, Li, Rahmati, Xiao,
  Prakash, Kohno, and Song]{eykholt2018robust}
K.~Eykholt, I.~Evtimov, E.~Fernandes, B.~Li, A.~Rahmati, C.~Xiao, A.~Prakash,
  T.~Kohno, and D.~Song.
\newblock Robust physical-world attacks on deep learning visual classification.
\newblock In \emph{Proceedings of the IEEE conference on computer vision and
  pattern recognition}, pages 1625--1634, 2018.

\bibitem[Hu et~al.(2021)Hu, Yang, Qiao, Zhao, and Wang]{hu2021seasondepth}
H.~Hu, B.~Yang, Z.~Qiao, D.~Zhao, and H.~Wang.
\newblock Seasondepth: Cross-season monocular depth prediction dataset and
  benchmark under multiple environments.
\newblock \emph{arXiv preprint arXiv:2011.04408}, 2021.

\bibitem[Yang and Carlone(2022)]{yang2022certifiably}
H.~Yang and L.~Carlone.
\newblock Certifiably optimal outlier-robust geometric perception: Semidefinite
  relaxations and scalable global optimization.
\newblock \emph{IEEE Transactions on Pattern Analysis and Machine
  Intelligence}, 2022.

\bibitem[Ferreira and Dias(2014)]{ferreira2014probabilistic}
J.~F. Ferreira and J.~M. Dias.
\newblock \emph{Probabilistic approaches to robotic perception}.
\newblock Springer, 2014.

\bibitem[Liu et~al.(2022)Liu, Amini, Takac, and Motee]{liu2022robustness}
G.~Liu, A.~Amini, M.~Takac, and N.~Motee.
\newblock Robustness analysis of classification using recurrent neural networks
  with perturbed sequential input.
\newblock \emph{arXiv preprint arXiv:2203.05403}, 2022.

\bibitem[Goodfellow et~al.(2014)Goodfellow, Shlens, and
  Szegedy]{goodfellow2014explaining}
I.~J. Goodfellow, J.~Shlens, and C.~Szegedy.
\newblock Explaining and harnessing adversarial examples.
\newblock \emph{arXiv preprint arXiv:1412.6572}, 2014.

\bibitem[Kurakin et~al.(2018)Kurakin, Goodfellow, and
  Bengio]{kurakin2018adversarial}
A.~Kurakin, I.~J. Goodfellow, and S.~Bengio.
\newblock Adversarial examples in the physical world.
\newblock In \emph{Artificial intelligence safety and security}, pages 99--112.
  Chapman and Hall/CRC, 2018.

\bibitem[Carlini and Wagner(2017)]{carlini2017towards}
N.~Carlini and D.~Wagner.
\newblock Towards evaluating the robustness of neural networks.
\newblock In \emph{2017 ieee symposium on security and privacy (sp)}, pages
  39--57. IEEE, 2017.

\bibitem[Xiao et~al.(2018)Xiao, Li, Zhu, He, Liu, and Song]{xiao2018generating}
C.~Xiao, B.~Li, J.~Y. Zhu, W.~He, M.~Liu, and D.~Song.
\newblock Generating adversarial examples with adversarial networks.
\newblock In \emph{27th International Joint Conference on Artificial
  Intelligence, IJCAI 2018}, pages 3905--3911. International Joint Conferences
  on Artificial Intelligence, 2018.

\bibitem[Liu et~al.(2019)Liu, Arief, and Zhao]{liu2019should}
Z.~Liu, M.~Arief, and D.~Zhao.
\newblock Where should we place lidars on the autonomous vehicle?-an optimal
  design approach.
\newblock In \emph{2019 International Conference on Robotics and Automation
  (ICRA)}, pages 2793--2799. IEEE, 2019.

\bibitem[Feng et~al.(2021)Feng, Lee, Durner, and Triebel]{feng2021bridging}
J.~Feng, J.~Lee, M.~Durner, and R.~Triebel.
\newblock Bridging the last mile in sim-to-real robot perception via bayesian
  active learning.
\newblock \emph{arXiv preprint arXiv:2109.11547}, 2021.

\bibitem[Hu et~al.(2022)Hu, Liu, Chitlangia, Agnihotri, and
  Zhao]{hu2022investigating}
H.~Hu, Z.~Liu, S.~Chitlangia, A.~Agnihotri, and D.~Zhao.
\newblock Investigating the impact of multi-lidar placement on object detection
  for autonomous driving.
\newblock In \emph{Proceedings of the IEEE/CVF Conference on Computer Vision
  and Pattern Recognition}, pages 2550--2559, 2022.

\bibitem[Tang et~al.(2017)Tang, Von~Gioi, Monasse, and
  Morel]{tang2017precision}
Z.~Tang, R.~G. Von~Gioi, P.~Monasse, and J.-M. Morel.
\newblock A precision analysis of camera distortion models.
\newblock \emph{IEEE Transactions on Image Processing}, 26\penalty0
  (6):\penalty0 2694--2704, 2017.

\bibitem[Tram{\`e}r et~al.(2018)Tram{\`e}r, Boneh, Kurakin, Goodfellow,
  Papernot, and McDaniel]{tramer2018ensemble}
F.~Tram{\`e}r, D.~Boneh, A.~Kurakin, I.~Goodfellow, N.~Papernot, and
  P.~McDaniel.
\newblock Ensemble adversarial training: Attacks and defenses.
\newblock In \emph{6th International Conference on Learning Representations,
  ICLR 2018-Conference Track Proceedings}, 2018.

\bibitem[Ma et~al.(2018)Ma, Li, Wang, Erfani, Wijewickrema, Schoenebeck, Song,
  Houle, and Bailey]{ma2018characterizing}
X.~Ma, B.~Li, Y.~Wang, S.~M. Erfani, S.~Wijewickrema, G.~Schoenebeck, D.~Song,
  M.~E. Houle, and J.~Bailey.
\newblock Characterizing adversarial subspaces using local intrinsic
  dimensionality.
\newblock In \emph{International Conference on Learning Representations}, 2018.

\bibitem[Tramer et~al.(2020)Tramer, Carlini, Brendel, and
  Madry]{tramer2020adaptive}
F.~Tramer, N.~Carlini, W.~Brendel, and A.~Madry.
\newblock On adaptive attacks to adversarial example defenses.
\newblock \emph{Advances in Neural Information Processing Systems},
  33:\penalty0 1633--1645, 2020.

\bibitem[Cohen et~al.(2019)Cohen, Rosenfeld, and Kolter]{cohen2019certified}
J.~Cohen, E.~Rosenfeld, and Z.~Kolter.
\newblock Certified adversarial robustness via randomized smoothing.
\newblock In \emph{International Conference on Machine Learning}, pages
  1310--1320. PMLR, 2019.

\bibitem[Tjeng et~al.(2018)Tjeng, Xiao, and Tedrake]{tjeng2018evaluating}
V.~Tjeng, K.~Y. Xiao, and R.~Tedrake.
\newblock Evaluating robustness of neural networks with mixed integer
  programming.
\newblock In \emph{International Conference on Learning Representations}, 2018.

\bibitem[Wong and Kolter(2018)]{wong2018provable}
E.~Wong and Z.~Kolter.
\newblock Provable defenses against adversarial examples via the convex outer
  adversarial polytope.
\newblock In \emph{International Conference on Machine Learning}, pages
  5286--5295. PMLR, 2018.

\bibitem[Li et~al.(2021)Li, Weber, Xu, Rimanic, Kailkhura, Xie, Zhang, and
  Li]{li2021tss}
L.~Li, M.~Weber, X.~Xu, L.~Rimanic, B.~Kailkhura, T.~Xie, C.~Zhang, and B.~Li.
\newblock Tss: Transformation-specific smoothing for robustness certification.
\newblock In \emph{Proceedings of the 2021 ACM SIGSAC Conference on Computer
  and Communications Security}, pages 535--557, 2021.

\bibitem[Ruoss et~al.(2021)Ruoss, Baader, Balunovi{\'c}, and
  Vechev]{ruoss2021efficient}
A.~Ruoss, M.~Baader, M.~Balunovi{\'c}, and M.~Vechev.
\newblock Efficient certification of spatial robustness.
\newblock In \emph{Proceedings of the AAAI Conference on Artificial
  Intelligence}, volume~35, pages 2504--2513, 2021.

\bibitem[Alfarra et~al.(2021)Alfarra, Bibi, Khan, Torr, and
  Ghanem]{alfarra2021deformrs}
M.~Alfarra, A.~Bibi, N.~Khan, P.~H. Torr, and B.~Ghanem.
\newblock Deformrs: Certifying input deformations with randomized smoothing.
\newblock \emph{arXiv preprint arXiv:2107.00996}, 2021.

\bibitem[Webots()]{Webots}
Webots.
\newblock http://www.cyberbotics.com.
\newblock URL \url{http://www.cyberbotics.com}.
\newblock Open-source Mobile Robot Simulation Software.

\bibitem[Hu et~al.(2021)Hu, Wang, Liu, and Chen]{hu2021domain}
H.~Hu, H.~Wang, Z.~Liu, and W.~Chen.
\newblock Domain-invariant similarity activation map contrastive learning for
  retrieval-based long-term visual localization.
\newblock \emph{IEEE/CAA Journal of Automatica Sinica}, 9\penalty0
  (2):\penalty0 313--328, 2021.

\bibitem[Hu et~al.(2019)Hu, Wang, Liu, Yang, Chen, and Xie]{hu2019retrieval}
H.~Hu, H.~Wang, Z.~Liu, C.~Yang, W.~Chen, and L.~Xie.
\newblock Retrieval-based localization based on domain-invariant feature
  learning under changing environments.
\newblock In \emph{2019 IEEE/RSJ International Conference on Intelligent Robots
  and Systems (IROS)}, pages 3684--3689. IEEE, 2019.

\bibitem[Tobin et~al.(2017)Tobin, Fong, Ray, Schneider, Zaremba, and
  Abbeel]{tobin2017domain}
J.~Tobin, R.~Fong, A.~Ray, J.~Schneider, W.~Zaremba, and P.~Abbeel.
\newblock Domain randomization for transferring deep neural networks from
  simulation to the real world.
\newblock In \emph{2017 IEEE/RSJ international conference on intelligent robots
  and systems (IROS)}, pages 23--30. IEEE, 2017.

\bibitem[Qiao et~al.(2021)Qiao, Hu, Shi, Chen, Liu, and
  Wang]{qiao2021registration}
Z.~Qiao, H.~Hu, W.~Shi, S.~Chen, Z.~Liu, and H.~Wang.
\newblock A registration-aided domain adaptation network for 3d point cloud
  based place recognition.
\newblock In \emph{2021 IEEE/RSJ International Conference on Intelligent Robots
  and Systems (IROS)}, pages 1317--1322. IEEE, 2021.

\bibitem[Li et~al.(2020)Li, Xie, and Li]{li2020sok}
L.~Li, T.~Xie, and B.~Li.
\newblock Sok: Certified robustness for deep neural networks.
\newblock \emph{arXiv preprint arXiv:2009.04131}, 2020.

\bibitem[Liu et~al.(2019)Liu, Arnon, Lazarus, Barrett, and
  Kochenderfer]{liu2019algorithms}
C.~Liu, T.~Arnon, C.~Lazarus, C.~Barrett, and M.~J. Kochenderfer.
\newblock Algorithms for verifying deep neural networks.
\newblock \emph{arXiv preprint arXiv:1903.06758}, 2019.

\bibitem[Madry et~al.(2018)Madry, Makelov, Schmidt, Tsipras, and
  Vladu]{madry2017towards}
A.~Madry, A.~Makelov, L.~Schmidt, D.~Tsipras, and A.~Vladu.
\newblock Towards deep learning models resistant to adversarial attacks.
\newblock In \emph{International Conference on Learning Representations}, 2018.

\bibitem[Zhang et~al.(2018)Zhang, Weng, Chen, Hsieh, and
  Daniel]{zhang2018efficient}
H.~Zhang, T.-W. Weng, P.-Y. Chen, C.-J. Hsieh, and L.~Daniel.
\newblock Efficient neural network robustness certification with general
  activation functions.
\newblock In \emph{Advances in neural information processing systems}, pages
  4939--4948, 2018.

\bibitem[Singh et~al.(2019)Singh, Gehr, P{\"u}schel, and
  Vechev]{singh2019abstract}
G.~Singh, T.~Gehr, M.~P{\"u}schel, and M.~Vechev.
\newblock An abstract domain for certifying neural networks.
\newblock \emph{Proceedings of the ACM on Programming Languages}, 3\penalty0
  (POPL):\penalty0 41, 2019.

\bibitem[Dathathri et~al.(2020)Dathathri, Dvijotham, Kurakin, Raghunathan,
  Uesato, Bunel, Shankar, Steinhardt, Goodfellow, Liang, and
  Kohli]{dathathri2020enabling}
S.~Dathathri, K.~Dvijotham, A.~Kurakin, A.~Raghunathan, J.~Uesato, R.~R. Bunel,
  S.~Shankar, J.~Steinhardt, I.~Goodfellow, P.~S. Liang, and P.~Kohli.
\newblock Enabling certification of verification-agnostic networks via
  memory-efficient semidefinite programming.
\newblock In H.~Larochelle, M.~Ranzato, R.~Hadsell, M.~F. Balcan, and H.~Lin,
  editors, \emph{Advances in Neural Information Processing Systems}, volume~33,
  pages 5318--5331, 2020.

\bibitem[Zhongkai et~al.(2022)Zhongkai, Ying, Dong, Su, and
  Zhu]{zhongkai2022gsmooth}
H.~Zhongkai, C.~Ying, Y.~Dong, H.~Su, and J.~Zhu.
\newblock {GS}mooth: Certified robustness against semantic transformations via
  generalized randomized smoothing.
\newblock In \emph{International Conference on Machine Learning}. PMLR, 2022.

\bibitem[Balunovi{\'c} et~al.(2019)Balunovi{\'c}, Baader, Singh, Gehr, and
  Vechev]{balunovic2019certifying}
M.~Balunovi{\'c}, M.~Baader, G.~Singh, T.~Gehr, and M.~Vechev.
\newblock Certifying geometric robustness of neural networks.
\newblock \emph{Advances in Neural Information Processing Systems 32}, 2019.

\bibitem[Mohapatra et~al.(2020)Mohapatra, Weng, Chen, Liu, and
  Daniel]{mohapatra2020towards}
J.~Mohapatra, T.-W. Weng, P.-Y. Chen, S.~Liu, and L.~Daniel.
\newblock Towards verifying robustness of neural networks against a family of
  semantic perturbations.
\newblock In \emph{Proceedings of the IEEE/CVF Conference on Computer Vision
  and Pattern Recognition}, pages 244--252, 2020.

\bibitem[Lorenz et~al.(2021)Lorenz, Ruoss, Balunovi{\'c}, Singh, and
  Vechev]{lorenz2021robustness}
T.~Lorenz, A.~Ruoss, M.~Balunovi{\'c}, G.~Singh, and M.~Vechev.
\newblock Robustness certification for point cloud models.
\newblock \emph{arXiv preprint arXiv:2103.16652}, 2021.

\bibitem[Fischer et~al.(2020)Fischer, Baader, and Vechev]{fischer2020certified}
M.~Fischer, M.~Baader, and M.~T. Vechev.
\newblock Certified defense to image transformations via randomized smoothing.
\newblock In \emph{NeurIPS}, 2020.

\bibitem[Chu et~al.(2022)Chu, Li, and Li]{chu2022tpc}
W.~Chu, L.~Li, and B.~Li.
\newblock Tpc: Transformation-specific smoothing for point cloud models.
\newblock In \emph{International Conference on Machine Learning}. PMLR, 2022.

\bibitem[Szeliski(2010)]{szeliski2010computer}
R.~Szeliski.
\newblock \emph{Computer vision: algorithms and applications}.
\newblock Springer Science \& Business Media, 2010.

\bibitem[Hao et~al.(2022)Hao, Ying, Dong, Su, Song, and Zhu]{hao2022gsmooth}
Z.~Hao, C.~Ying, Y.~Dong, H.~Su, J.~Song, and J.~Zhu.
\newblock Gsmooth: Certified robustness against semantic transformations via
  generalized randomized smoothing.
\newblock In \emph{International Conference on Machine Learning}, pages
  8465--8483. PMLR, 2022.

\bibitem[Engstrom et~al.(2019)Engstrom, Tran, Tsipras, Schmidt, and
  Madry]{engstrom2019exploring}
L.~Engstrom, B.~Tran, D.~Tsipras, L.~Schmidt, and A.~Madry.
\newblock Exploring the landscape of spatial robustness.
\newblock In \emph{International conference on machine learning}, pages
  1802--1811. PMLR, 2019.

\bibitem[Sitawarin et~al.(2022)Sitawarin, Golan-Strieb, and
  Wagner]{sitawarin2022demystifying}
C.~Sitawarin, Z.~J. Golan-Strieb, and D.~Wagner.
\newblock Demystifying the adversarial robustness of random transformation
  defenses.
\newblock In \emph{International Conference on Machine Learning}, pages
  20232--20252. PMLR, 2022.

\end{thebibliography}
\clearpage
\appendix{}
\section{Method Details and Proofs}
We first present the preliminary definitions in Section \ref{pre_defs}and provide details for Definition \ref{def:oracle_re} regarding translations and rotations on all 6 degrees of freedom in Section \ref{app:projections}. Then we show the proof for Lemma 1 and Theorem 1.
\subsection{Preliminary Definitions}
\label{pre_defs}

\begin{definition}[(\textbf{restated} of Definition \ref{def:oracle}) Position projective function]
\label{def:oracle_re}
For any  3D point $P=(X, Y, Z)\in\mathbb{P}\subset\mathbb{R}^3$ under the camera coordinate frame with  the camera intrinsic matrix $K$, based on the camera motion $\alpha= (\bm{\theta}, t) \in \mathcal{Z}\subset\mathbb{R}^6$ with  rotation matrix $ R = \exp(\bm{\theta}^\wedge) \in SO(3)$ and translation vector $t\in \mathbb{R}^3$, we define the position projective function $\rho: \mathbb{P}\times\mathcal{Z}\to \mathbb{R}^2$  and the depth function $D: \mathbb{P}\times\mathcal{Z}\to \mathbb{R}$  for point $P$ as
    \begin{align}
    \centering
    \label{def:projection_re}
    [\rho(P, \alpha), 1]^\top = \frac{1}{D(P, \alpha)}KR^{-1}(P - t), \quad
    D(P, \alpha) = [0, 0, 1] R^{-1}(P - t)
    \end{align}
\end{definition}

\begin{definition}[(\textbf{restated}  of Definition \ref{def:proj_trans}) Channel-wise projective transformation]
\label{def:proj_trans_re}
Given the position projection function $\rho: \mathbb{P}\times\mathcal{Z}\to \mathbb{R}^2$  and the depth function $D: \mathbb{P}\times \mathcal{Z}\to \mathbb{R}$ over dense 3D point cloud $\mathbb{P}$, define the  3D-2D global channel-wise projective transformation from $C$-channel colored point cloud $\mathbb{V}=(\mathbb{R}^C, \mathbb{P})\subset \mathbb{R}^{C+3}$ to $H\times W$ image gird $\mathcal{X}\subset \mathbb{R}^{C\times H\times W}$ as $O: \mathbb{V}\times\mathcal{Z} \to \mathcal{X}$ parameterized by camera motion $\alpha\in \mathcal{Z}$ using Floor function $\floor{\cdot}$,
\begin{align}
\label{def:min_pooling_re}
        x_{c,r,s}= O(V, \alpha)_{c,r,s} =         V_{c, P^*_\alpha}, \text{where } P^*_\alpha = \argmin \limits_{\{P\in \mathbb{P} \mid \floor{\rho(P,\alpha)} = (r,s) \}} D(P, \alpha)
\end{align}
Specifically, if $x = O(V, 0)$, we define the relative projective transformation  $\phi: \mathcal{X}\times\mathcal{Z}\to\mathcal{X}$ as,
\begin{align}
\label{phi_and_O_re}
\phi(x,\alpha)= O(V,\alpha).
\end{align}
\end{definition}

\begin{definition}[(\textbf{restated} of Definition \ref{def:smooth_parametric_classifier}) Camera motion $\varepsilon$-smoothed classifier]
\label{def:smooth_parametric_classifier_re}
Let $\phi: \mathcal{X}\times\mathcal{Z}\to\mathcal{X}$ be a relative projective transformation given the projected image $x$ at the origin of camera motion in the motion space $\mathcal{Z}$,
 and let $\varepsilon\sim\mathcal{P}_\varepsilon$ be a random camera motion taking values in $\mathcal{Z}$. Let $h: \mathcal{X}\to\mathcal{Y}$ be a base classifier $h(x) = \argmax_{y\in\mathcal{Y}} p(y\mid x)$, the expectation of projected image predictions $\phi(x, \varepsilon)$ over  camera motion distribution $\mathcal{P}_\varepsilon$  is $q(y\mid x;\varepsilon) := \mathbb{E}_{\varepsilon\sim\mathcal{P}_\varepsilon}p(y\mid\phi(x,\varepsilon))$. We define the $\varepsilon$-smoothed classifier $g:\mathcal{X}\to\mathcal{Y}$ as
     \begin{equation}
        g(x;\varepsilon) := \argmax_{y\in\mathcal{Y}} q(y\mid x;\varepsilon) = \argmax_{y\in\mathcal{Y}} \mathbb{E}_{\varepsilon\sim\mathcal{P}_\varepsilon} p(y\mid\phi(x,\varepsilon)).
    \end{equation}
\end{definition}

\subsection{Projective Function Details}
\label{app:projections}
Following \eqref{def:projection_re}, given 3D point  ${P}_0=(X_0, Y_0, Z_0)^T$ under the camera coordinate frame, with axis-angle or rotation vector $ \bm{\theta}=(\theta n_1, \theta n_2, \theta n_3)^T\in \mathbb{R}^3, \|\bm{\theta}\|_2=\theta\in\mathbb{R}$ and translation ${t}=(t_x, t_y, t_z)^T\in \mathbb{R}^3$, 
 the camera intrinsic matrix ${K}$ of the camera is shown below.
 $${K}=\begin{pmatrix} f_x & 0 & c_x \\ 0 & f_y &  c_y \\ 0 &0 & 1  \end{pmatrix}$$
  $$R^{-1}=\begin{pmatrix} \cos\theta+(1-\cos\theta)n_1^2 & (1-\cos\theta)n_1n_2+ n_3\sin\theta  & (1-\cos\theta)n_1n_3- n_2\sin\theta  \\ (1-\cos\theta)n_1n_2- n_3\sin\theta  & \cos\theta+(1-\cos\theta)n_2^2 &  (1-\cos\theta)n_2n_3+ n_1\sin\theta  \\ (1-\cos\theta)n_1n_3+ n_2\sin\theta  &(1-\cos\theta)n_2n_3- n_1\sin\theta  & \cos\theta+(1-\cos\theta)n_3^2  \end{pmatrix}$$
  First we find the depth of $P_0$ given camera pose $\alpha = \{{\theta},{t}\}$,
    \begin{align}
    \centering
    D(P_0, \alpha)&= [(1-\cos\theta)n_1n_3+ n_2\sin\theta ](X_0-t_x)\notag \\ &+ [(1-\cos\theta)n_2n_3- n_1\sin\theta ](Y_0-t_y) \notag \\ &+ [\cos\theta+(1-\cos\theta)n_3^2](Z_0-t_z) \notag
\end{align}
Then we find the pixel coordinates on the image.
  \begin{align}
    \centering
    \rho_1(P_0, \alpha)&= \frac{1}{D(P_0, \alpha)}\{[f_x [\cos\theta+(1-\cos\theta)n_1^2]+ c_x[(1-\cos\theta)n_1n_3+ n_2\sin\theta ]](X_0-t_x)\notag \\ &+ [f_x[(1-\cos\theta)n_1n_2+ n_3\sin\theta ] +c_x[(1-\cos\theta)n_2n_3- n_1\sin\theta ]](Y_0-t_y) \notag \\ &+ [f_x[(1-\cos\theta)n_1n_3- n_2\sin\theta ]+c_x[\cos\theta+(1-\cos\theta)n_3^2]](Z_0-t_z)\} \notag \\
        \rho_2(P_0, \alpha)&=  \frac{1}{D(P_0, \alpha)} \{ [f_y[\cos\theta+(1-\cos\theta)n_2^2] +c_y[(1-\cos\theta)n_2n_3- n_1\sin\theta ]](Y_0-t_y) \notag \\ &+ [f_y [(1-\cos\theta)n_1n_2- n_3\sin\theta ]+ c_y[(1-\cos\theta)n_1n_3+ n_2\sin\theta ]](X_0-t_x)  \notag \\ &+ [f_y[(1-\cos\theta)n_2n_3+ n_1\sin\theta ]+c_y[\cos\theta+(1-\cos\theta)n_3^2]](Z_0-t_z)\} \notag 
\end{align}

Specifically, the camera motion on each axis is shown as follows.
\subsubsection{$T_z$: {translation} along depth axis}
  In this case, we have $\theta=0, t_x = t_y = 0$
      \begin{align}
    \centering
    D_{T_z}(P_0, \alpha)=Z_0-t_z, \quad \notag
    \rho_{T_z}(P_0, \alpha)= (\frac{f_xX_0+c_x(Z_0-t_z)}{Z_0-t_z},
 \frac{f_yY_0+c_y(Z_0-t_z)}{Z_0-t_z})
\end{align}
  \subsubsection{$T_x$: {translation} along depth-orthogonal horizontal axis}
    In this case, we have $\theta=0, t_z = t_y = 0$
      \begin{align}
    \centering
    D_{T_x}(P_0, \alpha)=Z_0, \quad \notag
    \rho_{T_x}(P_0, \alpha)=(\frac{f_x(X_0-t_x)+c_xZ_0}{Z_0},
 \frac{f_yY_0+c_yZ_0}{Z_0})
\end{align}
        
        \subsubsection{$T_y$: {translation} along depth-orthogonal vertical axis}
    In this case, we have $\theta=0, t_z = t_x = 0$
      \begin{align}
    \centering
    D_{T_y}(P_0, \alpha)=Z_0, \quad \notag
    \rho_{T_y}(P_0, \alpha)= (\frac{f_xX_0+c_xZ_0}{Z_0},
\frac{f_y(Y_0-t_y)+c_yZ_0}{Z_0})
\end{align}
        
  \subsubsection{$R_z$: {rotation} around depth roll axis}
      In this case, we have $n_1=n_2=0, n_3 =1, t_x = t_y = t_z = 0$
      \begin{align}
    \centering
    D_{R_z}(P_0, \alpha)=Z_0, \quad \notag
    \rho_{R_z}(P_0, \alpha)= (\frac{f_x\cos\theta X_0+f_x \sin\theta  Y_0}{Z_0}+c_x,
\frac{f_y\cos\theta Y_0+f_y \sin\theta  X_0}{Z_0}+c_y)
\end{align}

    \subsubsection{$R_x$: {rotation} around depth-orthogonal pitch axis}
        In this case, we have $n_2=n_3=0, n_1 =1, t_x = t_y = t_z = 0$
      \begin{align}
    \centering
    D_{R_x}(P_0, \alpha)&=-\sin\theta Y_0+\cos\theta Z_0 \notag\\
    \rho_{R_x}(P_0, \alpha)&=(\frac{f_x X_0}{-Y_0\sin\theta +Z_0\cos\theta }+c_x,
     \notag
 \frac{Y_0\cos\theta +Z_0 \sin\theta}{-Y_0\sin\theta +Z_0\cos\theta }f_y +c_y) 
\end{align}

  \subsubsection{$R_y$: {rotation} around depth-orthogonal yaw axis}
        In this case, we have $n_1=n_3=0, n_2 =1, t_x = t_y = t_z = 0$
      \begin{align}
    \centering
    D_{R_y}(P_0, \alpha)&=\sin\theta X_0+\cos\theta Z_0 \notag
\\
    \rho_{R_y}(P_0, \alpha)&=(\frac{X_0\cos\theta- Z_0\sin\theta}{X_0\sin\theta + Z_0\cos\theta}f_x + c_x,\notag
 \frac{f_y Y_0}{X_0\sin\theta + Z_0\cos\theta} +c_y)
\end{align}


\subsection{Proof of Lemma \ref{lemma:resolvable_2d_proj}}
\begin{lemma}[\textbf{restated} of Lemma \ref{lemma:resolvable_2d_proj}, Compatible Relative Projection with Global Projection]
\label{lemma:resolvable_2d_proj_re}
            With a global projective transformation $O: \mathbb{V}\times\mathcal{Z} \to \mathcal{X}$ from 3D point cloud and a relative  projective  transformation $\phi: \mathcal{X}\times\mathcal{Z}\to\mathcal{X}$ given some original camera motions,  for any $\alpha_1\in\mathcal{Z}$ there exists an injective, continuously differentiable and non-vanishing-Jacobian function $\gamma_{\alpha_1}:\mathcal{Z}\to\mathcal{Z}$  such that 
            \begin{equation}
            \label{lemma:gamma_re}
                \phi(O(V, \alpha_1),\alpha_2) = O(V,\gamma_{\alpha_1}(\alpha_2)), V\in\mathbb{V},\alpha_2\in\mathcal{Z}.
            \end{equation}
\end{lemma}
\begin{proof}
	 Given the fixed colored point cloud map $V\in\mathbb{V}=(\mathbb{R}^C, \mathbb{P})$,
decompose the sequential relative camera motions $\alpha_1, \alpha_2$ into $R_1, t_1$ and $ R_2, t_2$, where $\alpha_1= (\theta_1, t_1), R_1 = \exp((\theta_1)^\wedge) \in SO(3)$ and $\alpha_2= (\theta_2, t_2), R_2 = \exp((\theta_2)^\wedge) \in SO(3)$. Following Definition \ref{def:oracle_re}, for any fixed 3D point $P_0\in \mathbb{P}$ under the initial camera pose, denote the coordinate after each relative camera motion as $P_1, P_2$, we have,
$$P_0 = R_1 P_1+t_1, P_1 = R_2 P_2+t_2$$
Therefore, the composed relative camera motion  is derived as,
$$P_0 = R_{1,2} P_2 + t_{1,2} =  R_1 R_2 P_2 + (R_1 t_2 + t_1)$$
	So the composition of camera motion is $\alpha_1 \circ \alpha_2 = (\theta_{1,2}, t_{1,2}) = (\theta_{1,2}, R_1 t_2 + t_1) $, where $ R_{1,2} = \exp((\theta_{1,2})^\wedge) = R_1 R_2$.
	Let the $\gamma$ function in \eqref{lemma:gamma_re}  be the composition of camera motion, i.e.,  $\gamma_{\alpha_1}(\alpha_2) = \alpha_1 \circ \alpha_2$, where the projection function with  min-pooling in Definition \ref{def:proj_trans_re} is also satisfied. Specifically,  if the rotation is around a fixed axis,  $\gamma_{\alpha_1}(\alpha_2) = \alpha_1 \circ \alpha_2 = \alpha_1 + \alpha_2$ holds due to the special case of multiplication in $SO(3)$. 
	
	Based on Definition \ref{def:proj_trans_re}, denote the point cloud coordinates after camera motion $\alpha_1$ to be $V^{\alpha_1}$, where the projected image is $$O(V, \alpha_1) = O(V^{\alpha_1}, 0) = x^{\alpha_1}$$
	 Then based on the composition of camera motion, it holds that
	$$O(V, \gamma_{\alpha_1}(\alpha_2)) = O(V, \alpha_1 \circ \alpha_2) = O(V^{\alpha_1}, \alpha_2)$$
	Following Equation \eqref{phi_and_O_re} in Definition \ref{def:proj_trans_re}, we have $\phi(x^{\alpha_1}, \alpha_2) = O(V^{\alpha_1}, \alpha_2)$
	So combining the derivations above, we have
	\begin{align*}
 \phi(O(V, \alpha_1),\alpha_2) & = \phi(O(V^{\alpha_1}, 0),\alpha_2) \\
 & = \phi(x^{\alpha_1},\alpha_2) \\
  & = O(V^{\alpha_1},\alpha_2) \\
   & = O(V, \alpha_1 \circ \alpha_2) \\
 & = O(V,\gamma_{\alpha_1}(\alpha_2))
	\end{align*}
which concludes the proof.
\end{proof}


\subsection{Proof of Theorem \ref{thm:resolvable_2d_proj}}

\begin{lemma}[Corollary 7 in \cite{li2021tss}, Corollary 3  in \cite{chu2022tpc}]
        \label{lem:gaussian_noise}
        Suppose $\mathcal{Z}=\mathbb{R}^m$, $\Sigma:=\mathrm{diag}(\sigma_1^2,\,\ldots,\sigma_m^2)$, $\varepsilon_0\sim\mathcal{N}(0,\,\Sigma)$ and $\varepsilon_1:=\alpha+\varepsilon_0$ for some $\alpha\in\mathbb{R}^m$.
        Suppose that $y_A = g(x;\varepsilon_0)$ at $x\in\mathcal{X}$ for some $y_A\in\mathcal{Y}$ and let $p_A,p_B\in[0,1]$ be bounds to the class probabilities, i.e.,
            \begin{equation}
            \label{papb_condition}
                q(y_A\mid x,\varepsilon_0) \geq p_A > p_B \geq \max_{y\neq y_A} q(y\mid x, \varepsilon_0).
            \end{equation}
        Then, it holds that $q(y_A\lvert\,x;\,\varepsilon_1) > \max_{y\neq y_A}q(y\lvert\,x;\,\varepsilon_1)$ if $\alpha$ satisfies
        \begin{equation}
            \label{eq:gaussian_condition}
            \sqrt{\sum_{i=1}^{m}\left(\dfrac{\alpha_i}{\sigma_i}\right)^2} < \dfrac{1}{2}\left(\Phi^{-1}(p_A) - \Phi^{-1}(p_B) \right).
        \end{equation}
    \end{lemma}
 The rigorous proof on Lemma\ref{lem:gaussian_noise} can be found in  \cite{li2021tss}.

\begin{theorem}[\textbf{restated} of Theorem \ref{thm:resolvable_2d_proj}, Robustness certification under camera motion with fixed-axis rotation]
\label{thm:resolvable_2d_proj_re}
Let $\alpha \in \mathcal{Z}\subset \mathbb{R}^6$ be  the parameters of projective transformation $\phi$ with translation $ (t_x, t_y, t_z)^T\in \mathbb{R}^3$ and fixed-axis rotation $ (\theta n_1, \theta n_2, \theta n_3)^T\in \mathbb{R}^3, \sum_{i=1}^3 n_i^2 = 1$,
suppose the composed camera motion $\varepsilon_1 \in \mathcal{Z}$  satisfies $ \phi(x, \varepsilon_1) =\phi(\phi(x, \varepsilon_0), \alpha) $ given some $\alpha \in \mathcal{Z}$ and zero-mean Gaussian motion $\varepsilon_0$ with variance $\sigma_x^2, \sigma_y^2, \sigma_z^2,\sigma_\theta^2$ for $t_x, t_y, t_z, \theta$ respectively,
        let $p_A,p_B\in[0,1]$ be bounds of the top-2 class probabilities for the motion smoothed model, i.e.,
            \begin{equation}
            \label{confidence_pab_re}
                q(y_A\mid x,\varepsilon_0) \geq p_A > p_B \geq \max_{y\neq y_A} q(y\mid x, \varepsilon_0).
                \end{equation}
        Then, it holds that $g(\phi(x, \alpha); \varepsilon_0) = g(x;\varepsilon_0)$ if $\alpha=(t_x, t_y, t_z,\theta n_1, \theta n_2, \theta n_3)^T$ satisfies
        \begin{equation}
        \label{equ:certify_condition_re}
            \sqrt{\left(\dfrac{\theta}{\sigma_\theta}\right)^2 + \left(\dfrac{t_x}{\sigma_x}\right)^2+\left(\dfrac{t_y}{\sigma_y}\right)^2+\left(\dfrac{t_z}{\sigma_z}\right)^2} < \dfrac{1}{2}\left(\Phi^{-1}(p_A) - \Phi^{-1}(p_B) \right).
        \end{equation}
\end{theorem}
\begin{proof}
For the original transformation $\phi$ parameterized with $\alpha=(t_x, t_y, t_z,\theta n_1, \theta n_2, \theta n_3)^T\in\mathcal{Z_\phi}\subset \mathbb{R}^6$ with the fixed normalized rotation axis $(n_1, n_2, n_3)$, we have the Gaussian noise for each entry $t_x\sim\mathcal{N}(0, \sigma_x),t_y\sim\mathcal{N}(0, \sigma_y),t_z\sim\mathcal{N}(0, \sigma_z),\theta n_1\sim\mathcal{N}(0, \sigma_\theta^2n_1^2),\theta n_2\sim\mathcal{N}(0, \sigma_\theta^2n_2^2),\theta n_3\sim\mathcal{N}(0, \sigma_\theta^2n_3^2)$.  We can find the covariance matrix $$\Sigma = \mathbb{E}[(\alpha-\mu_\alpha)(\alpha-\mu_\alpha)^\top]=\begin{pmatrix} \sigma_x^2 & 0 & 0 & 0 & 0 & 0 \\ 0 & \sigma_x^2 & 0 & 0 & 0 & 0  \\ 0 & 0 & \sigma_x^2 & 0 & 0 & 0 \\ 0 & 0 & 0 & n_1^2\sigma_\theta^2 & n_1 n_2\sigma_\theta^2 & n_1 n_3\sigma_\theta^2\\ 0 & 0 & 0 & n_1 n_2\sigma_\theta^2 & n_2^2\sigma_\theta^2 & n_2 n_3\sigma_\theta^2\\ 0 & 0 & 0 & n_1 n_3\sigma_\theta^2 & n_2 n_3\sigma_\theta^2 & n_3^2\sigma_\theta^2 \end{pmatrix}$$
Note that since the the last three entries regarding rotation angle $\theta$ is correlated, the covariance matrix is not full rank and not positive definite. Therefore, we can find the non-singular linear transformation $A$ to make all entries independent in  $\tilde\alpha = A\alpha\in\mathcal{\tilde Z_\phi}\subset \mathbb{R}^6$. Specifially, 
$$\tilde\alpha = A\alpha = \begin{pmatrix} 1 & 0 & 0 & 0 & 0 & 0 \\ 0 & 1 & 0 & 0 & 0 & 0  \\ 0 & 0 & 1 & 0 & 0 & 0 \\ 0 & 0 & 0 & n_1& n_2 &  n_3\\ 0 & 0 & 0 & n_1 & \frac{1}{n_2}-n_2 &  n_3\\ 0 & 0 & 0 & n_1  & n_2  & \frac{1}{n_3}-n_3 \end{pmatrix}\begin{pmatrix} t_x \\ t_y  \\ t_z \\ \theta n_1\\ \theta n_2\\ \theta n_3 \end{pmatrix} = \begin{pmatrix} t_x \\ t_y  \\ t_z \\ \theta \\ 0\\ 0 \end{pmatrix}$$
Then for $\tilde\alpha,\tilde\beta$ from the transformed parameter space where the transformation  is additive $\gamma_{\tilde\alpha}(\tilde\beta) = \tilde\alpha + \tilde\beta$, we find the covariance matrix as $$\tilde\Sigma = \mathbb{E}[(\tilde\alpha-\mu_{\tilde\alpha})(\tilde\alpha-\mu_{\tilde\alpha})^\top]=\mathrm{diag}(\sigma_x^2, \sigma_y^2, \sigma_z^2, \sigma_\theta^2, 0, 0)$$
For the projective transformation $\tilde\phi$ parameterized in space $\mathcal{\tilde Z_\phi}$, we have $$\tilde\phi(x, \tilde\varepsilon) = \tilde\phi(x, A\varepsilon)= \tilde O(V, A\varepsilon) = O(V, \varepsilon)= \phi(x, \varepsilon)$$. Therefore,  Lemma \ref{lemma:resolvable_2d_proj_re} holds for projective transformation $\tilde\phi$  over parameter space $\mathcal{\tilde Z_\phi}$. Therefore, for the composed transformation parameterized in space $\mathcal{\tilde Z_\phi}$ we have
\begin{align*}
    \centering
    \tilde\phi(x, \tilde\varepsilon_1) &=\tilde\phi(\tilde\phi(x, \tilde\varepsilon_0), \tilde\alpha) 
    =\tilde\phi(\tilde O(V, \tilde\varepsilon_0), \tilde\alpha) 
    =\tilde O(V, \gamma_{\tilde\varepsilon_0}(\tilde\alpha)) 
    =\tilde O(V, \tilde\varepsilon_0+\tilde\alpha) =\tilde \phi(x, \tilde\varepsilon_0+\tilde\alpha) \\
   \Longrightarrow \tilde\varepsilon_1& = \tilde\varepsilon_0+\tilde\alpha
    \end{align*}
 Together with $\mathcal{\tilde Z_\phi} = \mathbb{R}^6$, $\tilde\Sigma = \mathrm{diag}(\sigma_x^2, \sigma_y^2, \sigma_z^2, \sigma_\theta^2, 0, 0)$, we have $\tilde\varepsilon_0\sim\mathcal{N}(0,\,\tilde\Sigma)$ and $\tilde\varepsilon_1:=\tilde\alpha+\tilde\varepsilon_0$ for  $\tilde\alpha = (t_x, t_y, t_z,\theta, 0, 0)^T\in\mathbb{R}^6$,
$$\sqrt{\sum_{i=1}^{6}\left(\dfrac{\alpha_i}{\sigma_i}\right)^2} = \sqrt{\left(\dfrac{\theta}{\sigma_\theta}\right)^2 + \left(\dfrac{t_x}{\sigma_x}\right)^2+\left(\dfrac{t_y}{\sigma_y}\right)^2+\left(\dfrac{t_z}{\sigma_z}\right)^2}$$ 
the smoothed classifier for $p_A- p_B$-confidence condition of \eqref{papb_condition} in Lemma \ref{lem:gaussian_noise} is  satisfied based on ,
$$q(y\mid x;\tilde\varepsilon) =  \mathbb{E}_{\tilde\varepsilon\sim\mathcal{P}_{\tilde\varepsilon}} p(y\mid\tilde\phi(x,\tilde\varepsilon)) = \mathbb{E}_{\varepsilon\sim\mathcal{P}_{\varepsilon}} p(y\mid\tilde\phi(x,A\varepsilon)) =  \mathbb{E}_{\varepsilon\sim\mathcal{P}_{\varepsilon}} p(y\mid\phi(x,\varepsilon)) = q(y\mid x;\varepsilon) $$
so by Lemma \ref{lem:gaussian_noise}, it holds that 
$$q(y_A\lvert\,x;\,\varepsilon_1) = q(y_A\lvert\,x;\,\tilde\varepsilon_1) > \max_{y\neq y_A}q(y\lvert\,x;\,\tilde\varepsilon_1) = \max_{y\neq y_A}q(y\lvert\,x;\,\varepsilon_1) $$
Then according to Definition \ref{def:smooth_parametric_classifier_re}, we have 
\begin{align}
    \label{equ:same}
    g(x;\varepsilon_1) = \argmax_{y\in\mathcal{Y}} q(y\mid x;\varepsilon) = y_A = g(x;\tilde\varepsilon_0) =  g(x;\varepsilon_0)
\end{align}
Furthermore, combining Definition \ref{def:proj_trans_re}, Definition \ref{def:smooth_parametric_classifier_re} and Lemma \ref{lemma:resolvable_2d_proj_re}, it holds that
\begin{align*}
    g(\phi(x,\alpha);\varepsilon_0) &= g(O(V,\alpha);\varepsilon_0) & (\text{By Definition \ref{def:proj_trans_re}})\\
    &= \argmax_{y\in\mathcal{Y}} \mathbb{E}_{\varepsilon_0\sim\mathcal{P}_\varepsilon} p(y\mid\phi(O(V,\alpha),\varepsilon_0)) & (\text{By Definition \ref{def:smooth_parametric_classifier_re}})\\
    &=\argmax_{y\in\mathcal{Y}} \mathbb{E}_{\varepsilon_0\sim\mathcal{P}_\varepsilon} p(y\mid O(V, \gamma_\alpha(\varepsilon_0))) & (\text{By Lemma \ref{lemma:resolvable_2d_proj_re}})\\
    &=\argmax_{y\in\mathcal{Y}} \mathbb{E}_{\varepsilon_0\sim\mathcal{P}_\varepsilon} p(y\mid O(V, \varepsilon_1)) & \\
        &=\argmax_{y\in\mathcal{Y}} \mathbb{E}_{\varepsilon_0\sim\mathcal{P}_\varepsilon} p(y\mid \phi(x;\varepsilon_1)) &(\text{By Definition \ref{def:proj_trans_re}})\\
    &=g(x;\varepsilon_1) &(\text{By Definition \ref{def:smooth_parametric_classifier_re}})\\
& = g(x;\varepsilon_0) &(\text{By  Lemma \ref{lem:gaussian_noise} and Equation \eqref{equ:same}})
\end{align*}
which concludes the proof.
\end{proof}

\section{More Experiment Details}
\subsection{MetaRoom Dataset}
\begin{figure}[htbp]
	\centering
	
	\subfigure[Apple]{
		\begin{minipage}[b]{.18\linewidth}
			\centering
			\includegraphics[width=\textwidth]{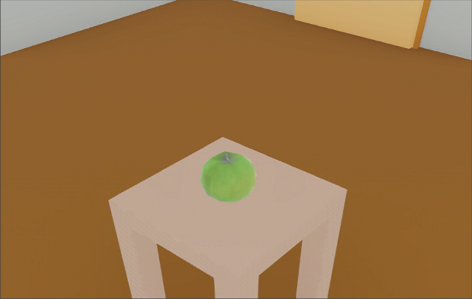}
		\end{minipage}
	}
	\subfigure[Beer Bottle]{
		\begin{minipage}[b]{.18\linewidth}
			\centering
			\includegraphics[width=\textwidth]{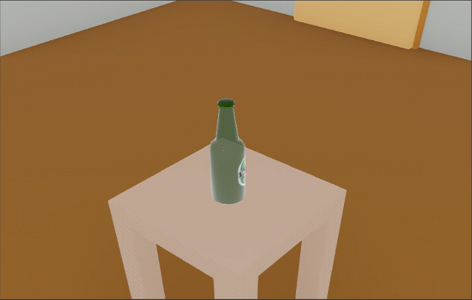}
		\end{minipage}
	}
	\subfigure[Biscuit Box]{
		\begin{minipage}[b]{.18\linewidth}
			\centering
			\includegraphics[width=\textwidth]{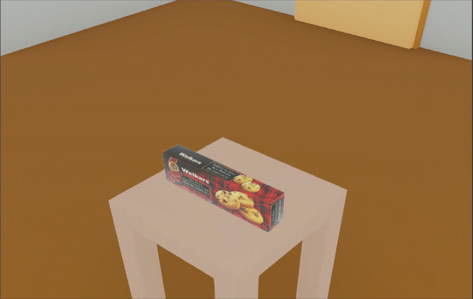}
		\end{minipage}
	}
	\subfigure[Book]{
		\begin{minipage}[b]{.18\linewidth}
			\centering
			\includegraphics[width=\textwidth]{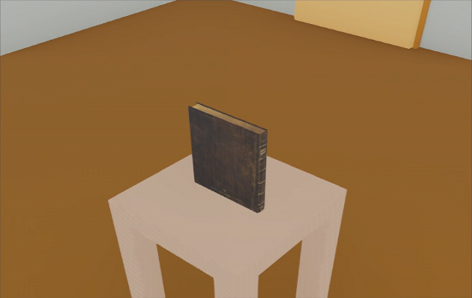}
		\end{minipage}
	}
	\subfigure[Can]{
	\begin{minipage}[b]{.18\linewidth}
		\centering
		\includegraphics[width=\textwidth]{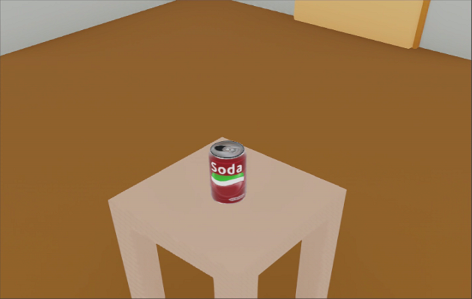}
	\end{minipage}
}
	\subfigure[Carafe]{
	\begin{minipage}[b]{.18\linewidth}
		\centering
		\includegraphics[width=\textwidth]{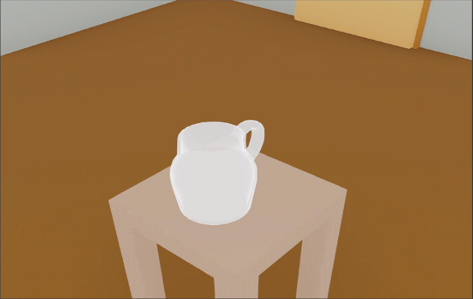}
	\end{minipage}
}
\subfigure[Cereal Box]{
	\begin{minipage}[b]{.18\linewidth}
		\centering
		\includegraphics[width=\textwidth]{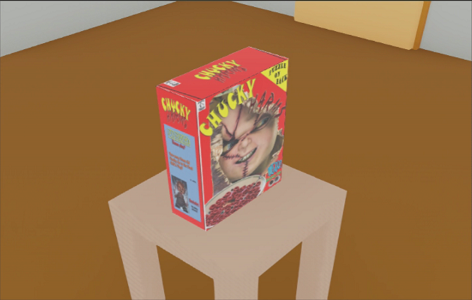}
	\end{minipage}
}
\subfigure[Fruit Bowl]{
	\begin{minipage}[b]{.18\linewidth}
		\centering
		\includegraphics[width=\textwidth]{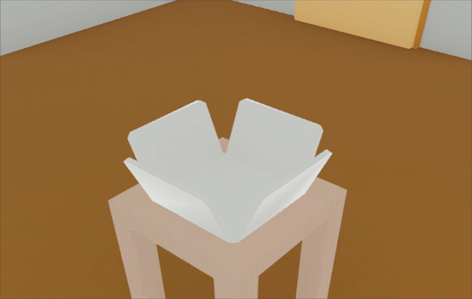}
	\end{minipage}
}
\subfigure[Honey Jar]{
	\begin{minipage}[b]{.18\linewidth}
		\centering
		\includegraphics[width=\textwidth]{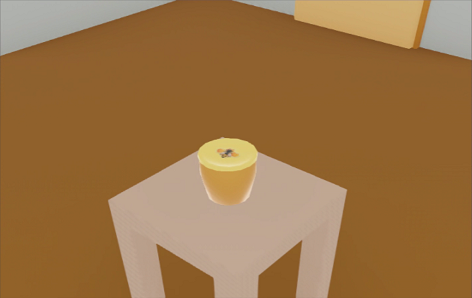}
	\end{minipage}
}
\subfigure[Jam Jar]{
	\begin{minipage}[b]{.18\linewidth}
		\centering
		\includegraphics[width=\textwidth]{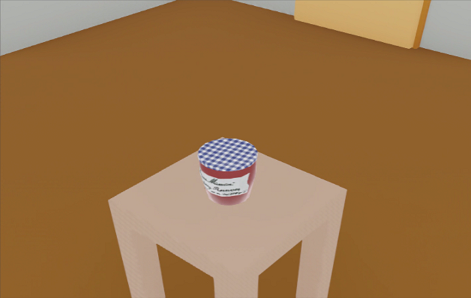}
	\end{minipage}
}
	\subfigure[Laptop]{
	\begin{minipage}[b]{.18\linewidth}
		\centering
		\includegraphics[width=\textwidth]{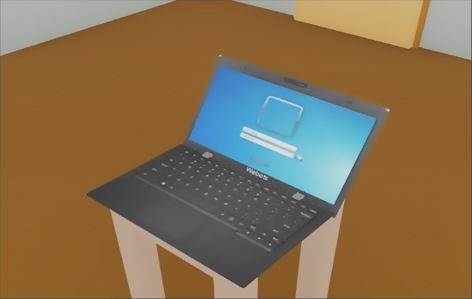}
	\end{minipage}
}
\subfigure[Monitor]{
	\begin{minipage}[b]{.18\linewidth}
		\centering
		\includegraphics[width=\textwidth]{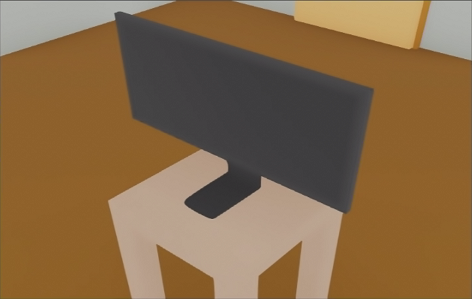}
	\end{minipage}
}
\subfigure[Office Telephone]{
	\begin{minipage}[b]{.18\linewidth}
		\centering
		\includegraphics[width=\textwidth]{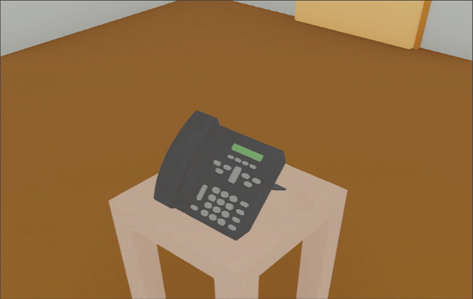}
	\end{minipage}
}
\subfigure[Orange]{
	\begin{minipage}[b]{.18\linewidth}
		\centering
		\includegraphics[width=\textwidth]{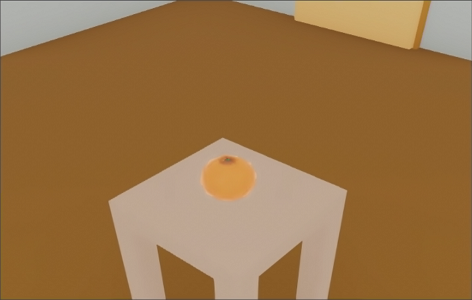}
	\end{minipage}
}
\subfigure[Paper Boat]{
	\begin{minipage}[b]{.18\linewidth}
		\centering
		\includegraphics[width=\textwidth]{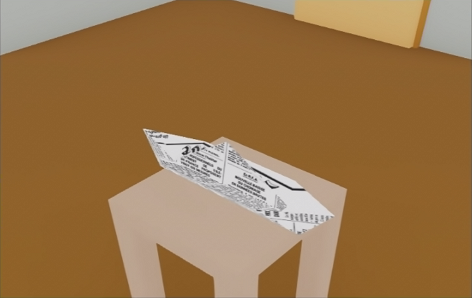}
	\end{minipage}
}
	\subfigure[Rubber Duck]{
	\begin{minipage}[b]{.18\linewidth}
		\centering
		\includegraphics[width=\textwidth]{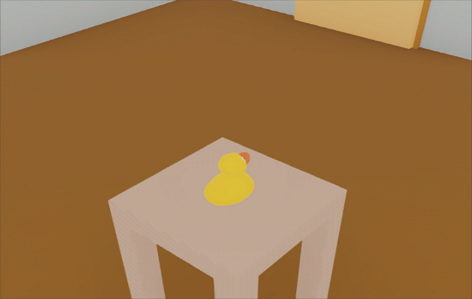}
	\end{minipage}
}
\subfigure[Soccer Ball]{
	\begin{minipage}[b]{.18\linewidth}
		\centering
		\includegraphics[width=\textwidth]{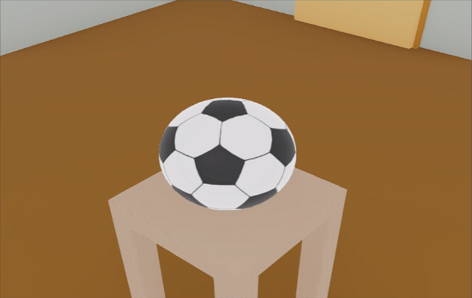}
	\end{minipage}
}
\subfigure[Telephone]{
	\begin{minipage}[b]{.18\linewidth}
		\centering
		\includegraphics[width=\textwidth]{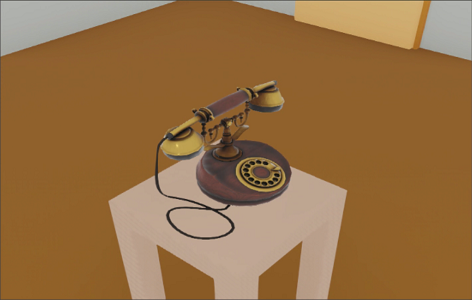}
	\end{minipage}
}
\subfigure[Water Bottle]{
	\begin{minipage}[b]{.18\linewidth}
		\centering
		\includegraphics[width=\textwidth]{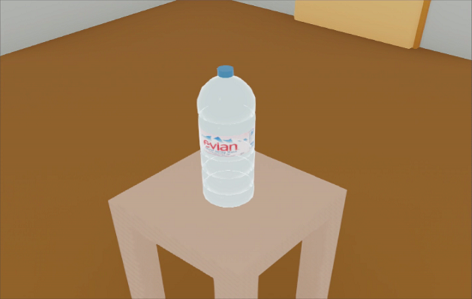}
	\end{minipage}
}
\subfigure[Wine Glass]{
	\begin{minipage}[b]{.18\linewidth}
		\centering
		\includegraphics[width=\textwidth]{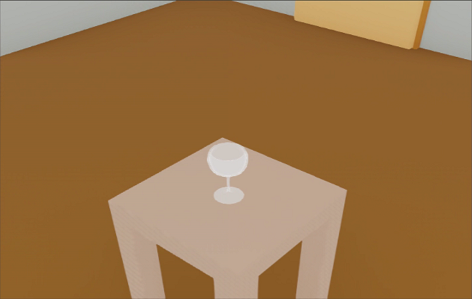}
	\end{minipage}
}
	\caption{All the objects in the MetaRoom dataset}
	\label{fig:all_objects}
\end{figure}

The entire room contains four surrounding walls with a length of 6 m and a height of 3.7 m, a ceiling and floor with size of 6 m $\times$ 6 m. On the walls, there is a door with the default size, a window with the default size, two paintings, two photographs, two closets, a blackboard, and a clock. In the center of the room, there is a small four-leg table with the size of 0.5 m $\times$ 0.5 m and the height of 1 m. In order to make the reconstructed point cloud have a consistent appearance under different camera perspectives, the texture of all the walls, ceiling, floor, door, window, and the table is \textit{Roughcast} to avoid reflections. All the objects are listed in Figure \ref{fig:all_objects}.

To collect the whole point cloud, we first collect point cloud maps for table objects, background walls, floor, and ceiling, respectively, and then merge them together with downsampling to the density of $0.0025$ m. To collect each point cloud map, we rotate the camera around the object where the focal length of the camera is $3090.194$ 
and resolution is $1280\times 720$. We collect the camera poses and images with the traverses under roll angle of $-5 ^\circ, 0^\circ, 5^\circ$, yaw angle of  $0 ^\circ \sim 360^\circ$ with the interval of $10^\circ$, pitch angle between $30^\circ\sim 80^\circ$ with the interval of $5^\circ$ and radius of 2.1 m, 2.5 m, 3.0 m, 3.4 m for table object, background walls, floor and ceiling with corresponding camera orientations. See the attached code for more details.

For the collection of the training set and test set, we use the collection strategy stated in Section 4.1 of the main text. To speed up the image projection for testing and evaluation,  we first collect images with poses using focal length of $3090.194$ 
and resolution of  $1280\times 720$ to construct the local point cloud within each perturbation radius. Then collect the test camera poses with the intrinsic matrix of the focal length of $386.274$
and resolution of  $160\times 90$, where the low-resolution images are directly used for model training and evaluation without resizing to rigorously follow  Definition \ref{def:proj_trans_re}. For model training with data augmentation, we collect all images under perturbations offline for computational efficiency, which is consistent with training epochs of undefended vanilla models.

\begin{table}
	\centering
\begin{tabular}{|c|c|c|c|c|c|c|}
\hline
Camera motion type                                                                & $T_z$ & $T_x$ & $T_y$ & $R_z$                                    & $R_x$                                                 & $R_y$                                                 \\ \hline
\begin{tabular}[c]{@{}c@{}}Motion Agumentation\\ (Gaussian variance)\end{tabular} & 0.1 m         & 0.05 m        & 0.05 m        & \begin{tabular}[c]{@{}c@{}}0.122 rad\\ (7$^\circ$)\end{tabular} & \begin{tabular}[c]{@{}c@{}}0.0436 rad\\ (2.5$^\circ$)\end{tabular} & \begin{tabular}[c]{@{}c@{}}0.0436 rad\\ (2.5$^\circ$)\end{tabular} \\ \hline
\end{tabular}
\vspace{2mm}
	\caption{Motion augmentation details}
	\label{tab:aug}
\end{table}

\subsection{Model Training}
To train the base classifiers, we train the ResNet-18 and ResNet-50 models from random initialization for both motions augmented and undefended vanilla models. The motion augmentation for each training sample is implemented with perturbations under Gaussian distribution   and  $\sigma$ for each axis is shown in Figure \ref{tab:aug}.  The inputs are without resizing or other image-based augmented  for a clean and fair comparison, followed by channel normalization to 0.5 mean and variance. The models are trained with a batch size of 32 and a learning rate of 0.001 for 100 epochs. We remark that the goal of the model training is to ensure that the base classifiers can perform normally on the test set without overfitting or underfitting and we focus more on the evaluation for robustness analysis, so we did not fully explore the training potential. The performance of course can be further improved through tuning the hyper parameters and model architectures in more effective ways.  All the experiments are conducted on NVIDIA A6000 with 48G GPU and 128G RAM.

\subsection{Evaluation Details}

\begin{table}
	\centering
\begin{tabular}{|c|c|c|c|c|}
	\hline
	\begin{tabular}[c]{@{}c@{}}Training strategy\\ (defense or not)\end{tabular}  & \begin{tabular}[c]{@{}c@{}}Evaluation strategy\\ (smoothing or not)\end{tabular} & \begin{tabular}[c]{@{}c@{}}Benign Acc\\ (in Main Text)\end{tabular} & \begin{tabular}[c]{@{}c@{}}Emp. Robust Acc\\ (in Main Text)\end{tabular} & \begin{tabular}[c]{@{}c@{}}Certified Acc\\ (in Main Text)\end{tabular} \\ \hline
	\multirow{2}{*}{\begin{tabular}[c]{@{}c@{}}Undefended\\ Vinilla\end{tabular}} & Base Model                                                                       & Table \ref{tab:base_smoothed_vanilla}                                                             & Table \ref{tab:base_smoothed_vanilla}                                                                  &  ---                                                                      \\ \cline{2-5} 
	& Motion Smoothed                                                                  & Table \ref{tab:vanilla_certify}, \ref{tab:base_smoothed_vanilla}                                                          & Table \ref{tab:vanilla_certify}, \ref{tab:base_smoothed_vanilla}                                                              & Table \ref{tab:vanilla_certify}                                                                \\ \hline
	\multirow{2}{*}{\begin{tabular}[c]{@{}c@{}}Motion\\ Augmented\end{tabular}}   & Base Model                                                                       & Table \ref{tab:base_smoothed}                                                             & Table \ref{tab:base_smoothed}                                                                  &    ---                                                                    \\ \cline{2-5} 
	& Motion Smoothed                                                                  & Table \ref{tab:base_smoothed}, \ref{tab:vanilla_certify}                                                          & Table \ref{tab:base_smoothed}, \ref{tab:vanilla_certify}                                                               & Table \ref{tab:vanilla_certify}                                                                \\ \hline
\end{tabular}
\vspace{2mm}
	\caption{Instruction and clarification in the evaluation for different models}
	\label{tab:all}
\end{table}
We first list all the experimental results in Table \ref{tab:all} to give instructions to locate the performance for different models under training and evaluation strategies.

\textbf{Benign and empirical robust accuracy for base models.}
To evaluate base models, we only use the metrics of benign accuracy and empirical robust accuracy because certified accuracy cannot be obtained without smoothing. The base classifiers are the trained models and we test them directly on the test set for benign accuracy while we use offline motion perturbed images under uniform distribution around each test sample to obtain the worst-case empirical robust accuracy. 

\textbf{Benign, empirical robust accuracy for smoothed models.} For the smoothing model, since the smoothing is with Gaussian distribution through online Monte Carlo sampling for each test image, we adopt online perturbation within a certain radius given the point cloud and camera pose in the test set to obtain the benign and empirical accuracy for a fair comparison. For the empirical accuracy through Monte Carlo, we adopt 100 samples to obtain top-2 classes with the confidence of $99\%$ and batch size of 100 for the smoothed classifier following \cite{cohen2019certified}. 

\textcolor{revise}{\textbf{Comparison between 5 and 100 perturbed empirical robust accuracy.} }
\label{details}
\textcolor{revise}{From Table \ref{100_perturb}, it can be seen that grid search based attack 100-perturbed attacks are just a little bit stronger than 5-perturbed ones, so either can be used to approximate the worst-case adversarial samples although the gradient-based attack cannot be implemented directly in the camera motion transformation space \cite{engstrom2019exploring, sitawarin2022demystifying}. Note that  the gap between 5-perturbed and 100-perturbed attacks is less for motion augmented  models compared to undefended vanilla models, showing that motion augmentation can improve empirical robustness against uniform perturbation.}

\textbf{Certified accuracy for smoothed models.} For the certification of the smoothed model, we also use Monte Carlo sampling for smoothing over confidence of $99\%$ using 1000 samples and 100 samples to obtain the top-1 classes with a batch size of 200. To make the projection more efficient during smoothing, we use the local dense point cloud map reconstructed from maximum perturbations on each axis. Specifically, we adopt two-stage down-sampling strategy of \texttt{uniform\_down\_sample} and \texttt{voxel\_down\_sample}. The down-sampling hyperparameters are listed in Table \ref{tab:projection_hyper}.
\begin{table}[]
	\centering
\begin{tabular}{|c|c|c|}
	\hline
	Camera motion type & \begin{tabular}[c]{@{}c@{}}Uniform Down Sample\\ (every k points)\end{tabular} & \begin{tabular}[c]{@{}c@{}}Uniform Down Sample\\ (density: m)\end{tabular} \\ \hline
	Translation Z      & 7                                                                              & 0.0133                                                                     \\ \hline
	Translation X      & 6                                                                              & 0.01365                                                                    \\ \hline
	Translation Y      & 7                                                                              & 0.0137                                                                     \\ \hline
	Rotation Z         & 7                                                                              & 0.0135                                                                     \\ \hline
	Rotation X         & 6                                                                              & 0.01355                                                                    \\ \hline
	Rotation Y         & 7                                                                              & 0.0134                                                                     \\ \hline
\end{tabular}
\vspace{2mm}
	\caption{Hyperparameters for two-stage down sampling to speed up smoothing}
\label{tab:projection_hyper}
\end{table}

\begin{table}[]
	\centering
\begin{tabular}{ccc}
\hline
Camera Motion Types                                   & Smoothed ResNet18               & Smoothed ResNet50                  \\ \hline
$T_z$, radius [-0.1m, 0.1m]               & Vanilla / Motion Aug.           & Vanilla / Motion Aug.              \\ \hline
5-perturbed Emp. Robust Acc.                          & 0.817    /      0.833           & 0.617    / 0.850                   \\
\textcolor{revise}{100-perturbed Emp. Robust Acc.}                        & \textcolor{revise}{0.783}     /       \textcolor{revise}{0.817}         & \textcolor{revise}{0.567}    /      \textcolor{revise}{0.825}              \\ \hline
$T_x$, radius [-0.05m, 0.05m]             & Vanilla / Motion Aug.           & Vanilla / Motion Aug.              \\ \hline
5-perturbed Emp. Robust Acc.                          & 0.783           /    0.875      & 0.675    /     0.825               \\
\textcolor{revise}{100-perturbed Emp. Robust Acc.}                           & \textcolor{revise}{0.758}        /       \textcolor{revise}{0.867}      & \textcolor{revise}{0.617}   /    \textcolor{revise}{0.800}                \\ \hline
$T_y$, radius [-0.05m, 0.05m]             & Vanilla / Motion Aug.           & Vanilla / Motion Aug.              \\ \hline
5-perturbed Emp. Robust Acc.                          & 0.825            /     0.875    & 0.767                  /     0.925 \\
\textcolor{revise}{100-perturbed Emp. Robust Acc.}                           & \textcolor{revise}{0.792}          /          \textcolor{revise}{0.842} & \textcolor{revise}{0.758}        / \textcolor{revise}{0.908}               \\ \hline
$R_z$, radius [-7$^\circ$, 7$^\circ$]     & Vanilla / Motion Aug.           & Vanilla / Motion Aug.              \\ \hline
5-perturbed Emp. Robust Acc.                          & 0.742      /     0.933          & 0.717               /     0.917    \\
\textcolor{revise}{100-perturbed Emp. Robust Acc.}                           & \textcolor{revise}{0.717}        /           \textcolor{revise}{0.892}  & \textcolor{revise}{0.675}    /  \textcolor{revise}{0.917}                  \\ \hline
$R_x$, radius [-2.5$^\circ$, 2.5$^\circ$] & Vanilla / Motion Aug.           & Vanilla / Motion Aug.              \\ \hline
5-perturbed Emp. Robust Acc.                          & 0.800      /       0.942        & 0.742                  /     0.933 \\
\textcolor{revise}{100-perturbed Emp. Robust Acc.}                           & \textcolor{revise}{0.750}   /       \textcolor{revise}{0.892}           & \textcolor{revise}{0.692}        /      \textcolor{revise}{0.917}          \\ \hline
$R_y$, radius [-2.5$^\circ$, 2.5$^\circ$] & Vanilla / Motion Aug.           & Vanilla / Motion Aug.              \\ \hline
5-perturbed Emp. Robust Acc.                          & 0.875         /        0.925    & 0.783                /   0.992     \\
\textcolor{revise}{100-perturbed Emp. Robust Acc.}                           & \textcolor{revise}{0.808}         /       \textcolor{revise}{0.925}     & \textcolor{revise}{0.742}   /   \textcolor{revise}{0.983}                  \\ \hline
\end{tabular}
\vspace{2mm}
	\caption{\textcolor{revise}{Comparison of performance in terms of 5 and 100 perturbed empirical robust accuracy.}}
\label{100_perturb}
\end{table}

\textbf{Smoothed v.s. base vanilla model.} For the undefended vanilla models,  Tab. \ref{tab:base_smoothed_vanilla} presents that smoothed models have higher empirical robust  accuracy and the gap between benign and empirical robust accuracy becomes less after motion smoothing compared to the base models, showing that smoothing strategy works not only for well-defended motion augmented models, but also 
 for undefended vanilla models. Compared to Tab. \ref{tab:base_smoothed}, there is more robustness/accuracy trade-off in rotation for the vanilla models, which implies that  motion augmentation as a defense in model training  helps to improve the robustness better against rotational perturbations than translational perturbations.

\begin{table}[]
\centering
\begin{tabular}{ccc}
\hline
Camera Motion Types &  Vanilla ResNet18 &  Vanilla ResNet50  \\
\hline
$T_z$, radius {[}-0.1m, 0.1m{]} &  Base / Smoothed &  Base / Smoothed \\
\hline
Benign Accuracy                       &                0.800    /   \textbf{0.858}                &               \textbf{0.708}       /     0.675                   \\
\textcolor{revise}{100-perturbed Emp. Robust Acc.}             &               \textcolor{revise}{0.708}    /    \textcolor{revise}{\textbf{0.783}}                  &  \textcolor{revise}{0.517}        /      \textcolor{revise}{\textbf{0.567}}       \\
\hline
$T_x$, radius {[}-0.05m, 0.05m{]} &  Base / Smoothed &  Base / Smoothed \\
\hline
Benign Accuracy                       &        0.817            /   \textbf{0.825}                &                  0.717       /    \textbf{0.767}                    \\
\textcolor{revise}{100-perturbed Emp. Robust Acc.}               &         \textcolor{revise}{0.608}          /    \textcolor{revise}{\textbf{0.758}}                 &    \textcolor{revise}{0.467}       /       \textcolor{revise}{\textbf{0.617}}       \\
\hline
$T_y$, radius {[}-0.05m, 0.05m{]} &  Base / Smoothed &  Base / Smoothed \\
\hline
Benign Accuracy                       &      0.825              /     \textbf{0.850}               &                 \textbf{0.817}       /     0.792                   \\
\textcolor{revise}{100-perturbed Emp. Robust Acc.}               &         \textcolor{revise}{0.675}          /    \textcolor{revise}{\textbf{0.792}}               &     \textcolor{revise}{0.674}      /    \textcolor{revise}{\textbf{0.758}}         \\
\hline
$R_z$, radius {[}-7$^\circ$, 7$^\circ${]} &  Base / Smoothed &  Base / Smoothed \\
\hline
Benign Accuracy                       &           0.800        /   \textbf{0.817}               &                   \textbf{0.783}     /   0.758                     \\
\textcolor{revise}{100-perturbed Emp. Robust Acc.}               &           \textcolor{revise}{0.667}         /     \textcolor{revise}{\textbf{0.717}}              &   \textcolor{revise}{0.558}         /   \textcolor{revise}{\textbf{0.675}}           \\
\hline
$R_x$, radius {[}-2.5$^\circ$, 2.5$^\circ${]} &  Base / Smoothed &  Base / Smoothed \\
\hline
Benign Accuracy                       &            \textbf{0.867}        /   0.842                &                       \textbf{0.767}      /    \textbf{0.767}                    \\
\textcolor{revise}{100-perturbed Emp. Robust Acc.}               &             \textcolor{revise}{0.667}       /     \textcolor{revise}{\textbf{0.750}}               &    \textcolor{revise}{0.467}      / \textcolor{revise}{\textbf{0.692}}                 \\
\hline
$R_y$, radius {[}-2.5$^\circ$, 2.5$^\circ${]} &  Base / Smoothed &  Base / Smoothed \\
\hline
Benign Accuracy                       &         \textbf{0.917}          /   0.892                 &                  0.800     /    \textbf{0.808}                  \\
\textcolor{revise}{100-perturbed Emp. Robust Acc.}               &           \textcolor{revise}{0.692}         /      \textcolor{revise}{\textbf{0.808}}              &          \textcolor{revise}{0.600}      / \textcolor{revise}{\textbf{0.742}}    \\
\hline
\end{tabular}
\vspace{1mm}
\caption{\small The comparison between base vanilla models and  smoothed  vanilla models in benign and \textcolor{revise}{100-perturbed empirical robust  accuracy} for  all camera motions. The higher one between  each base  and  smoothed  model is in \textbf{bold}.}
\label{tab:base_smoothed_vanilla}
\vspace{-8mm}
\end{table}

\begin{figure}[htbp]
	\centering
	
	\subfigure[Diamond]{
		\begin{minipage}[b]{.3\linewidth}
			\centering
			\includegraphics[width=\textwidth]{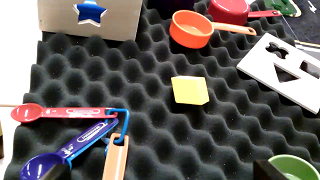}
		\end{minipage}
	}
	\subfigure[Oval]{
		\begin{minipage}[b]{.3\linewidth}
			\centering
			\includegraphics[width=\textwidth]{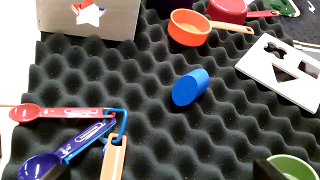}
		\end{minipage}
	}
	\subfigure[Quatrefoil]{
		\begin{minipage}[b]{.3\linewidth}
			\centering
			\includegraphics[width=\textwidth]{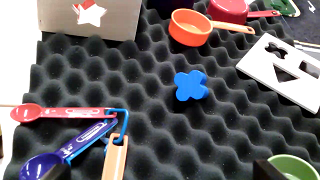}
		\end{minipage}
	}
	\subfigure[Rectangle]{
		\begin{minipage}[b]{.3\linewidth}
			\centering
			\includegraphics[width=\textwidth]{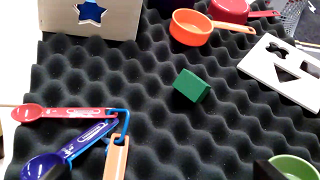}
		\end{minipage}
	}
	\subfigure[Star]{
	\begin{minipage}[b]{.3\linewidth}
		\centering
		\includegraphics[width=\textwidth]{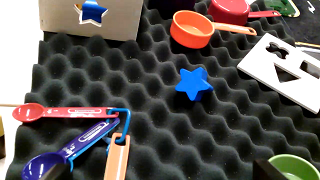}
	\end{minipage}
}
	\subfigure[Trapezoid]{
	\begin{minipage}[b]{.3\linewidth}
		\centering
		\includegraphics[width=\textwidth]{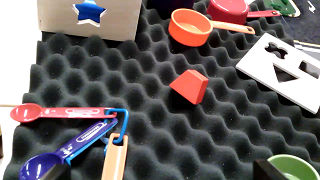}
	\end{minipage}
}
	\caption{All the objects in the real-world robot experiment}
	\label{fig:real_images}
\end{figure}

\textcolor{revise}{\subsection{Real-world Robot Experiment Details}}
\label{sec:real_robot}
\textcolor{revise}{The working zone of the pick-place environment is on the table with the size of $1m \times 1m$. The objects used for the perception model can be seen in Figure \ref{fig:real_images}.
Since we do not do any defense or augmentation in model training, the results in the Table \ref{tab:real_robot} are from the vanilla model. The gap to avoid the overlapping between the test set and training set is  $20^\circ$ for roll angle, $5^\circ$ for pitch angle, $10^\circ$ for yaw angle, and $1.6cm$ for radius. To make the robot application more practical, we remark that the smoothing method is used to improve empirical robustness so we omit the benign accuracy using the smoothing method, which can be found in Table \ref{tab:base_smoothed} and \ref{tab:base_smoothed_vanilla}.}

\end{document}